\documentclass[letterpaper]{article} 
\usepackage{aaai24}  
\usepackage{times}  
\usepackage{helvet}  
\usepackage{courier}  
\usepackage[hyphens]{url}  
\usepackage{graphicx} 
\usepackage{natbib}  
\usepackage{caption} 
\newcommand{\namemodel}{\textsc{COMBHelper}{}}

\usepackage{algorithm}
\usepackage{algpseudocodex}

\usepackage{amsfonts} 
\usepackage{amsmath}
\usepackage{cases}
\usepackage{xcolor}
\usepackage{booktabs}
\usepackage{multirow}
\usepackage{makecell}
\usepackage{subcaption}

\usepackage{newfloat}
\usepackage{listings}
\DeclareCaptionStyle{ruled}{labelfont=normalfont,labelsep=colon,strut=off} 
\floatstyle{ruled}
\newfloat{listing}{tb}{lst}{}
\floatname{listing}{Listing}
\title{\namemodel: A Neural Approach to Reduce Search Space for Graph Combinatorial Problems}
\author{
    Hao Tian\textsuperscript{\rm 1},
    Sourav Medya\textsuperscript{\rm 2},
    Wei Ye\textsuperscript{\rm 1}
}
\affiliations{
    \textsuperscript{\rm 1}College of Electronic and Information Engineering, Tongji University, Shanghai, China \\
    \textsuperscript{\rm 2}Department of Computer Science, University of Illinois Chicago, IL, USA \\

    \{2133036, yew\}@tongji.edu.cn, medya@uic.edu
}

\usepackage{bibentry}

\begin{document}

\maketitle

\begin{abstract}

Combinatorial Optimization (CO) problems over graphs appear routinely in many applications such as in optimizing traffic, viral marketing in social networks, and matching for job allocation. Due to their combinatorial nature, these problems are often NP-hard. Existing approximation algorithms and heuristics rely on the search space to find the solutions and become time-consuming when this space is large. In this paper, we design a neural method called {\namemodel} to reduce this space and thus improve the efficiency of the traditional CO algorithms based on node selection. Specifically, it employs a Graph Neural Network (GNN) to identify promising nodes for the solution set. This pruned search space is then fed to the traditional CO algorithms. {\namemodel} also uses a Knowledge Distillation (KD) module and a problem-specific boosting module to bring further efficiency and efficacy. Our extensive experiments show that the traditional CO algorithms with {\namemodel} are at least 2 times faster than their original versions. Our code is available at Github\footnote{\url{https://github.com/1041877801/COMBHelper}}.

\end{abstract}

\section{Introduction}

CO problems over graphs arise in many applications such as social networks \cite{chaoji2012recommendations}, transportation \cite{james2019online}, health-care \cite{wilder2018optimizing}, and biology \cite{hossain2020automated}. Due to their combinatorial nature, these problems are often NP-hard, and thus the design of optimal polynomial-time algorithms is infeasible. Traditional CO algorithms such as approximation algorithms and heuristics \cite{hochba1997approximation,papadimitriou1998combinatorial,vazirani2001approximation,williamson2011design} are usually used to solve such problems in practice. Approximation algorithms take polynomial time and provide theoretical bounds on the quality of the solutions. On the other hand, the design of heuristics lacks provable guarantees and requires empirical expertise in specific CO problems. Although some of these algorithms \cite{andrade2012fast,mathieu2017study,hoos2004stochastic} can find (nearly) optimal solutions or approximate high-quality solutions, they have one common bottleneck in terms of \textit{efficiency}. Since the efficiency of generating solutions relies on the size of the search space, it becomes time-consuming to obtain the solution when the size of the search space is large.

In recent years, machine learning (ML) methods, especially neural approaches, have been applied to design effective heuristics for CO problems \cite{khalil2017learning,li2018combinatorial,manchanda2020gcomb,barrett2020exploratory}. These neural approaches boost the performance of heuristics by exploiting the expressiveness of GNNs \cite{kipf2016semi,hamilton2017inductive,gilmer2020message} and generate high-quality solutions in practice. However, they often consist of complex models with a large number of parameters. They predict the probability of each node being included in the final solution and add the best node (i.e., the node with the largest probability) to the current solution. Such process is repeated until the final solutions are obtained. Thus, they can still suffer from the \textit{efficiency} issue of the search space being large.

To address the \textit{efficiency} issue, an intuitive idea is to reduce the size of the search space for the CO problems. For example, \cite{grassia2019learning,lauri2020learning,fitzpatrick2021learning} view the CO problem as a classification task. They train a linear classifier such as logistic regression \cite{cox1958regression} to identify bad candidates (nodes or edges) unlikely to be included in the solution and prune them from the search space. In our work, we build upon such idea and propose a novel framework called {\namemodel} to address the \textit{efficiency} issue of existing CO algorithms based on \textit{node selection}. Instead of training a linear model for classification, we adopt a non-linear neural model, i.e., a GNN model, which fully exploits the graph structural information and has shown remarkable performance on \textit{node classification} tasks. In particular, we train a GNN model to identify the \textit{good candidates} (i.e., nodes) that are likely to be included in the solution and prune the \textit{bad or unlikely ones} from the search space. Moreover, we apply two modules to further enhance the GNN model. The first one is a KD module used to compress the GNN model and reduce the inference time. The other one is a problem-specific boosting module, which is designed to improve the performance of node classification so that the search space for specific CO problems is pruned more precisely. Afterwards, we execute traditional CO algorithms on the reduced search space, which can accelerate the process of finding the solution set. We briefly summarize our contributions as follows:
\begin{itemize}
	\item We propose a novel framework called {\namemodel} to improve the efficiency of existing graph CO algorithms based on node selection by reducing the search space with a GNN-based module.
	
	\item We adopt the KD framework and design a problem-specific boosting module for the GNN model to further improve the efficiency and efficacy of {\namemodel}.
	
	\item Extensive experiments demonstrate the efficiency and efficacy of {\namemodel} on both synthetic and real-world datasets. In particular, traditional CO algorithms with {\namemodel} are at least 2 times faster than their original versions.
\end{itemize}

\section{Related Work}

\subsubsection{Neural Approaches on CO Problems.}

Neural CO approaches have been applied to design more effective heuristics and have shown significant performance on CO problems \cite{vinyals2015pointer,bello2016neural,khalil2017learning,li2018combinatorial,ahn2020learning}. Ptr-Net \cite{vinyals2015pointer} is an improved Recurrent Neural Network(RNN) model whose output length is not fixed. At each iteration of solution generation, the decoder uses the attention mechanisms \cite{bahdanau2014neural} to calculate a probability for each element and add the best one (with the largest probability) into the solution. \cite{bello2016neural} proposes a framework to solve the Traveling Salesman Problem (TSP). It trains a Ptr-Net \cite{vinyals2015pointer} by reinforcement learning, which is used to sequentially generate solutions. The network parameters are optimized by an actor-critic algorithm that combines two different policy gradient methods. S2V-DQN \cite{khalil2017learning} solves three CO problems: Minimum Vertex Cover (MVC), Maximum Cut (MAXCUT) and TSP. It trains a deep Q-network (DQN), which is parameterized by a GNN called structure2vec \cite{dai2016discriminative}, to construct solutions in a greedy manner. \cite{li2018combinatorial} combines Graph Convolutional Network (GCN) \cite{kipf2016semi} with a tree search module to solve NP-hard problems. GCN is trained to generate multiple likelihoods for each node and a set of potential solutions is obtained via the tree search procedure. Then the best one is selected as the final solution. 
LwD \cite{ahn2020learning} proposes a deep reinforcement learning framework to solve the Maximum Independent Set (MIS) problem on large graphs. The agent iteratively makes or defers a decision (add one node into the independent set) until all the nodes are determined.

\subsubsection{Pruning Search Space for CO Problems.}

To improve the efficiency of existing CO algorithms, some CO frameworks \cite{grassia2019learning,lauri2020learning,manchanda2020gcomb,fitzpatrick2021learning} prune the search space of the CO problems and just pay attention to the reduced search space. \cite{grassia2019learning,lauri2020learning} solve the Maximum Clique Enumeration (MCE) problem and design a multi-stage pruning strategy. In each stage, they learn a new classifier with hand-crafted features that contain both graph-theoretic and statistical features. GCOMB \cite{manchanda2020gcomb} is proposed to solve budget-constrained CO problems such as Maximum Coverage Problem (MCP) and Influence Maximization (IM). GCOMB trains a GCN model \cite{hamilton2017inductive} to prune bad nodes and a Q-learning network is trained on the pruned search space to predict the solution sequentially. \cite{fitzpatrick2021learning} proposes a framework to solve the TSP, which presents the edges with three different kinds of features and trains an edge-based classifier to prune edges unlikely to be included in the optimal solution.

\subsubsection{Knowledge Distillation on GNNs.}

The concept of KD is first introduced in \cite{hinton2015distilling} whose goal is to transfer the knowledge from a high-capacity teacher model to a simple student model without loss of validity. Recently, KD has been widely applied on GNNs \cite{yang2020distilling,yan2020tinygnn,deng2021graph,zhang2021graph,guo2023boosting}. \cite{yang2020distilling,yan2020tinygnn} adopts KD to compress GNN models while preserving the local structure information, which achieves competitive performance on node classification. GFKD \cite{deng2021graph} is a KD framework for graph-level tasks, which does not need observable graph data. GLNN \cite{zhang2021graph} aims to distill knowledge from GNNs to Multi-Layer Perceptrons (MLPs), which accelerates the model inference process and performs well on node-level tasks. BGNN \cite{guo2023boosting} designs two novel modules to enhance GNNs: adaptive temperature and weight boosting, which shows superior performance on both node-level and graph-level tasks.

\section{Problem Formulation}

In this paper, we focus on graph CO problems that appear routinely in multiple domains such as in biology \cite{hossain2020automated} and scheduling \cite{bansal2010inapproximability}. In particular, we apply our neural framework on two CO problems over graphs. Let $\mathcal{G}=(\mathcal{V},\mathcal{E})$ denote an undirected graph, where $\mathcal{V}$ and $\mathcal{E}$ are the node set and the edge set respectively. The problems are as follows:
\begin{itemize}
	\item \textbf{Minimum Vertex Cover (MVC):} Given an undirected graph $\mathcal{G}$, find a subset of nodes $\mathcal{S}\subseteq\mathcal{V}$ such that each edge ${e}\in\mathcal{E}$ in the graph is adjacent to at least one node in subset $\mathcal{S}$, and the size of $\mathcal{S}$ is minimized.
	
	\item \textbf{Maximum Independent Set (MIS):} Given an undirected graph $\mathcal{G}$, find a subset of nodes $\mathcal{S}\subseteq\mathcal{V}$ such that no two nodes in $\mathcal{S}$ are connected by an edge ${e}\in\mathcal{E}$ in the graph, and the size of $\mathcal{S}$ is maximized.
\end{itemize}

\subsubsection{Our Objective.}

Given a CO problem over graphs, our main objective is to prune the search space of the problem via a neural model and perform traditional CO algorithms on the reduced search space to obtain the final solution. We showcase the performance of our method on three well-known CO algorithms: linear programming \cite{land2010automatic}, greedy algorithm \cite{mathieu2017study,khalil2017learning} and local search \cite{hoos2004stochastic,andrade2012fast}. These CO algorithms are illustrated in more detail in the supplementary.

\section{Our Proposed Method: \namemodel}

\begin{figure*}[t]
	\centering
	\includegraphics[width=0.725\textwidth]{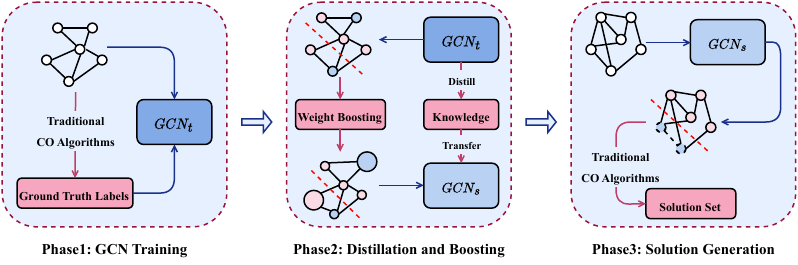}
	\caption{Overview of the proposed {\namemodel}. In phase 1, we generate the ground truth node labels with traditional CO algorithms on a small graph and train a teacher GCN (GCN$_{t}$) in a supervised manner; In phase 2, we train a student GCN (GCN$_{s}$) with two modules: (1) the KD module which transfers the knowledge from GCN$_{t}$ to GCN$_{s}$ and (2) the problem-specific boosting module which is used to amplify the weights of the nodes misclassified by GCN$_{t}$; In phase 3, we use GCN$_{s}$ to prune the search space of traditional CO algorithms on a large graph, i.e., predict \textit{good nodes} (nodes predicted as candidate elements in the solution). Then we perform traditional CO algorithms on the reduced search space to obtain the final solution.}
	\label{fig:frameworkoverview}
\end{figure*}

As mentioned above, most of the existing CO algorithms suffer from efficiency issues due to large search space. To mitigate this issue, we propose a framework called {\namemodel} (illustrated in Figure \ref{fig:frameworkoverview}) to prune the search space for CO problems and hence accelerate the solution generation process. Given an undirected graph $\mathcal{G}$, our goal is to predict the importance of each node and find the nodes that are likely to be included in the final solution. To achieve this, we apply a GNN-based architecture, {\namemodel}, to identify the quality of the nodes. For simplicity, we exemplify by GCN \cite{kipf2016semi} but any other model can be used. {\namemodel} has three major steps and each step is discussed separately in the following sections:
\begin{itemize}
	\item \textbf{Classification with GCN:} We train a GCN model to identify promising nodes that are likely to be included in the solution and subsequently, prune the search space. 
	
	\item \textbf{Knowledge Distillation and Problem-Specific Boosting:} To further improve the performance of the trained GCN model, we adopt the KD strategy during the training procedure and design a problem-specific boosting module for different CO problems.
	
	\item \textbf{Final Solution Generation:} Lastly, to generate the final solution, we again apply the traditional CO algorithms on the reduced search space.
\end{itemize}

\subsection{Classification with GCN}

First, we view the problem as a classification task where we classify the nodes based on their potential to be included in the solution set. The GCN model is trained in a supervised manner on the nodes with their ground truth labels. We use three traditional CO algorithms to generate the labels: (1) Linear Programming (LP), (2) Greedy algorithm (GD), and (3) Local Search (LS). Here, we show how to use LP to generate ground truth labels, which are often used to achieve optimal solutions for CO problems.

\subsubsection{Generate Labels.} 

We take the MVC problem defined in the previous section as an example: for a given graph $\mathcal{G}=(\mathcal{V},\mathcal{E})$, each node $v\in\mathcal{V}$ is denoted as a binary variable $x_{v}$. Then the LP problem can be formulated as follow:

\begin{align}
	\label{eqn:mvclpobj} \min&\sum_{v\in\mathcal{V}}x_{v} \\
	\label{eqn:mvclpconstraint1} &x_{v}\in\{0,1\} \\
	\label{eqn:mvclpconstraint2} &x_{v}+x_{u}\ge1,(v,u)\in\mathcal{E}
\end{align}
where Equation \ref{eqn:mvclpobj} is the objective of MVC that minimizes the number of nodes in the solution; Equation \ref{eqn:mvclpconstraint1} means the value of each node variable is selected as 0 or 1, where $x_{v}=1$ means node $v$ is included in the solution and $x_{v}=0$ otherwise; Equation \ref{eqn:mvclpconstraint2} makes sure that at least one node is included in the solution for each edge in the graph. After solving the LP problem, we obtain the final solution $\mathcal{S}=\{v|x_{v}=1\}$ that helps us to create the one-hot labels for each node in the graph, i.e., $\left[1,0\right]$ for $v\notin\mathcal{S}$ and $\left[0,1\right]$ for $v\in\mathcal{S}$.

\subsubsection{GCN Training.}

We describe the training procedure for GCN as follows. Assume $\mathcal{D}=(\mathcal{G},\mathbf{Y})$ is a training instance, where $\mathcal{G}=(\mathcal{V},\mathcal{E})$ is the input graph and $\mathbf{Y}\in\mathbb{R}^{|\mathcal{V}| \times 2}$ is the ground truth label generated by LP, i.e. $\mathbf{y}_{v}$ is the one-hot label of node $v$. $\mathcal{G}$ is also associated with a feature matrix $\mathbf{X}\in\mathbb{R}^{|\mathcal{V}| \times d}$, where the $i$-th row of $\mathbf{X}$ represents a $d$-dimensional feature vector of node $v_{i}$. Let $f(\mathcal{G};\Theta)$ denote the GCN parameterized by $\Theta$ and its forward propagation process is illustrated as follows:

We initialize the input embedding of each node $v$ as $h_{v}^{(0)}=\mathbf{x}_{v}$ and the embedding in the $(k+1)$-th layer is computed from that in the $k$-th layer as follow:
\begin{equation}
	\label{eqn:gcn}
	h_{v}^{(k+1)}=\sigma(h_{v}^{(k)}\theta_{1}^{k+1}+\sum_{u\in\mathcal{N}(v)}h_{u}^{(k)}\theta_{2}^{k+1})
\end{equation}
where $\mathcal{N}(v)$ is a set that contains the neighboring nodes of $v$, $\sigma(\cdot)$ is the ReLU \cite{nair2010rectified} activation function, $\theta_{1}^{k+1}\in\mathbb{R}^{d^{(k)} \times d^{(k+1)}}$, $\theta_{2}^{k+1}\in\mathbb{R}^{d^{(k)} \times d^{(k+1)}}$ are trainable parameters and $d^{(k)}$, $d^{(k+1)}$ are the dimensions of the $k$-th layer and $(k+1)$-th layer, respectively.

Our objective is to minimize the cross-entropy loss during the supervised training process, which is as follow:
\begin{equation}
	\mathcal{L}_{label}=-\sum_{v\in\mathcal{V}_{train}}\textbf{y}_{v}\log(\hat{\textbf{y}}_{v})
\end{equation}
where $\mathcal{V}_{train}$ is the set of training nodes, $\mathbf{y}_{v}$ is the label of node $v$, $\hat{\textbf{y}}_{v}=$ softmax($z_{v}$) is the prediction and $z_{v}=f_{v}(\mathcal{G};\Theta)$ is the output logits of node $v$, where the logits is a non-normalized probability vector, which is always passed to a normalization function such as  softmax($\cdot$) or sigmoid($\cdot$).

\subsection{Knowledge Distillation \& Boosting}

To further improve the efficiency and efficacy of {\namemodel}, we apply a KD method on the trained GCN. More specifically, we integrate a problem-specific weight boosting module into the KD framework in order to compress the GCN model and improve its performance.

\subsubsection{Knowledge Distillation.}

KD is used to reduce the number of parameters in the trained GCN model and improve the efficiency of {\namemodel}. We adapt the KD method introduced in \cite{hinton2015distilling}, which is widely used in many frameworks. The main idea is to transfer the knowledge learned by a larger teacher model to a simpler or smaller student model without loss of quality. In {\namemodel}, the trained GCN is the teacher model, which learns knowledge about the nodes from the traditional CO algorithms such as LP on the small graphs. Its knowledge is transferred to a student GCN model by minimizing the cross-entropy loss between its output logits ($z_{v}^{t}$) and the output logits ($z_{v}^{s}$) of the student GCN. This can be formulated as follows:
\begin{equation}
	\mathcal{L}_{KD}=-\sum_{v\in\mathcal{V}_{train}}{\hat{\textbf{y}}_{v}^{t}\log(\hat{\textbf{y}}_{v}^{s})}
\end{equation}
where $\hat{\textbf{y}}_{v}^{t}=$ softmax($z_{v}^{t}/T$) and $\hat{\textbf{y}}_{v}^{s}=$ softmax($z_{v}^{s}/T$) are the predictions of the teacher GCN and the student GCN respectively, and $T$ is the temperature parameter used to soften the output logits of GCN.
\subsubsection{Boosting Module.}

In this study, we consider two different CO problems. To increase the efficacy of {\namemodel}, we design an additional problem-specific boosting module. This module improves the supervised training process of the student GCN. Such boosting module has been used in a different context to improve the training of the student model \cite{guo2023boosting}. The main idea is to amplify the weights of the nodes that are wrongly classified by the teacher GCN. This helps the student model to pay more attention to those nodes and to make a more precise prediction. We apply the weight updating function in the Adaboost algorithm \cite{freund1999short,sun2019adagcn} to update the node weights since our task is binary node classification. For each node $v\in\mathcal{V}_{train}$, its node weight is initialized as $w_{v}=\frac{1}{|\mathcal{V}_{train}|}$ and updated as follow:
\begin{numcases}{w_{v}=}
	w_{v}\cdot\exp(-\frac{1}{2}\ln\frac{1-\epsilon}{\epsilon}) \label{eqn:true} \\
	w_{v}\cdot\exp(\frac{1}{2}\ln\frac{1-\epsilon}{\epsilon}) \label{eqn:false}
\end{numcases}
where $\epsilon$ is the error rate of the teacher GCN. The node weight is updated by Equation \ref{eqn:true} if it is classified correctly and updated by Equation \ref{eqn:false} if it is misclassified.

\subsubsection{Problem-Specific Design.}

To make our weight-boosting function problem-specific, we incorporate graph structural information such as node degree. In general, nodes with high degrees are more likely to be included in the solution set of the MVC problem, and nodes with low degrees are more likely to be included in the solution set of the MIS problem. We exploit this intuition and let the student GCN focus more on the nodes with high degrees while addressing the MVC problem and on the ones with low degrees for the MIS problem. The problem-specific weight function can be formulated as follows:
\begin{align}
	\label{eqn:mvcweight} w_{v}&=w_{v}\cdot\mbox{norm}(deg_{v})\qquad\quad\mbox{(For MVC)}\\
	\label{eqn:misweight} w_{v}&=w_{v}\cdot\mbox{norm}(1/deg_{v})\qquad\mbox{(For MIS)}
\end{align}
where $deg_{v}$ is the degree of node $v$, $\mbox{norm}(\cdot)$ is the normalization function, e.g., $\mbox{norm}(deg_{v})=\frac{deg_{v}}{\sum_{v\in\mathcal{V}_{train}}deg_{v}}$.

\subsubsection{Overall Objective.}

The loss function of our framework {\namemodel} consists of two parts: the KD loss and the supervised training loss. They are combined with parameter $\lambda$, which is used to control the balance between them:
\begin{equation}\small
	\begin{split}
		\mathcal{L}
		& =\lambda\mathcal{L}_{KD}+(1-\lambda)\mathcal{L}_{label} \\
		& =-\lambda \sum_{v\in\mathcal{V}_{train}}{\hat{\textbf{y}}_{v}^{t}\log(\hat{\textbf{y}}_{v}^{s})}-(1-\lambda)\sum_{v\in\mathcal{V}_{train}}w_{v}\textbf{y}_{v}\log(\hat{\textbf{y}}_{v}^{s})
	\end{split}
\end{equation}

\subsection{Solution Generation}

The last step of {\namemodel} is to generate the final solution. Firstly, we use the trained student GCN to assign a binary label for each node. Nodes predicted as label 1 (defined as \textit{good nodes}) will be reserved in the search space since they are more likely to be included in the final solution set. Afterwards, we perform the traditional CO algorithms on the reduced search space (i.e., \textit{the set of good nodes}) to obtain the final solution.

We demonstrate the MVC problem as an example and perform LP on the \textit{good nodes} (denote as $\mathcal{V}_{g}$). Then the LP problem can be formulated as:
\begin{align}
	\label{eqn:ourmvclpobj} \min&\sum_{v\in\mathcal{V}_{g}}x_{v} \\
	\label{eqn:ourmvclpconstraint1} &x_{v}\in\{0,1\}\\
	\label{eqn:ourmvclpconstraint2} &x_{v}+x_{u}\ge1,(v,u)\in\mathcal{E},v\in\mathcal{V}_{g}~\mbox{or}~u\in\mathcal{V}_{g}
\end{align}
where Equation \ref{eqn:ourmvclpobj} is the modified objective that minimizes the number of \textit{good nodes} in the solution; Equation \ref{eqn:ourmvclpconstraint1} means the value of each \textit{good node} variable is selected from 0 and 1, where $x_{v}=1$ means node $v$ is included in the final solution and $x_{v}=0$ otherwise; Equation \ref{eqn:ourmvclpconstraint2} ensures that at least one \textit{good node} is included in the solution for each edge in the graph.

Besides LP, we also perform GD and LS on \textit{the set of good nodes} to obtain the final solution for the MVC problem. In addition, we generate solutions for the MIS problem with all these three traditional CO algorithms (Please see the supplementary for more details).

\section{Experimental Evaluation}

\subsection{Experimental Setup}

\subsubsection{Datasets.}
Both synthetic and real-world datasets are used in our experiments and they are introduced as follows: 
\begin{itemize}
	\item \textbf{Synthetic datasets:} We use Barab{\'a}si-Albert (BA) graph \cite{albert2002statistical}, which can model many real-world networks. We set the edge density to 4 and use BA-$N$ to denote the BA graph with $N$ nodes.
	\item \textbf{Real-world datasets:} The statistics of the datasets used in our experiments are shown in Table \ref{tab:realworld}. Pubmed \cite{sen2008collective} and DBLP \cite{pan2016tri} are citation networks and other datasets are SNAP networks \cite{snapnets}. All the networks are undirected.
\end{itemize}

\begin{table}[!htp]
	\small
	\centering
	\begin{tabular}{l|cc}
		\toprule
		\textbf{Dataset} & \textbf{Nodes \#} & \textbf{Edges \#} \\
		\midrule
		Alpha & 3.8K & 24.2K \\
		OTC & 5.9K & 35.6K \\
		DBLP & 17.7K & 52.9K \\
		Pubmed & 19.7K & 44.3K \\
		Brightkite & 58.2K & 214.1K \\
		Slashdot0811 & 77.4K & 905.5K \\
		Slashdot0922 & 82.2K & 948.5K \\
		Gowalla & 196.6K & 950.3K \\
		\bottomrule
	\end{tabular}
	\caption{Statistics of real-world datasets.}
	\label{tab:realworld}
\end{table}

\subsubsection{Baselines.}

We select three traditional CO algorithms (mentioned in the Problem Formulation section) as the baselines: (1) linear programming (LP), (2) greedy algorithm (GD), and (3) local search (LS). For each baseline, we compare it with two improved versions: {Baseline+\namemodel$_{pt}$} ({\namemodel$_{pt}$} means {\namemodel} without KD and boosting modules, where $pt$ stands for \textbf{p}re-trained \textbf{t}eacher) and {Baseline+\namemodel}.

\subsubsection{Training and Testing.}

For synthetic datasets, {\namemodel} is trained on BA graphs with 1000 nodes (BA-1K) and tested on BA graphs with $n$ nodes, where $n$ ranges from \{5K, 10K, 20K, 50K, 100K\}. For real-world datasets, {\namemodel} is trained on Cora \cite{sen2008collective} and tested on the datasets listed in Table \ref{tab:realworld}. More details are discussed in the supplementary.

\subsubsection{Evaluation Metrics.}

(1) To evaluate the efficiency of {\namemodel}, we compare the running time of each baseline with its improved versions (with {\namemodel}). (2) In addition, we use the average speed-up to reflect the efficiency of {\namemodel}, which is calculated by $speedup=\frac{time_{B}}{time_{I}}$, where $time_{B}$ and $time_{I}$ are the running times of the \textbf{B}aseline and its \textbf{I}mproved version (with {\namemodel}) respectively. (3) To evaluate the efficacy or quality of {\namemodel}, we report the solution size on the MIS problem (the larger the better). For the MVC problem, we report the solution size with the coverage (since we cannot cover all the edges, see the supplementary for more details). The coverage (the higher the better) is calculated by $coverage=\frac{|\mathcal{E}^{*}|}{|\mathcal{E}|}$, where $|\mathcal{E}^{*}|$ is the number of edges covered by the solution.

\subsection{Results}

\textit{We show that our proposed method produces better or comparable results (results of synthetic datasets are in the supplementary) while being much faster in all settings.}

\subsubsection{Performance on LP.}

All the LP-based solutions are obtained by CPLEX 12.9 \cite{nickel2021ibm} and we set the time limit to 1 hour.
Table \ref{tab:mvcresultreallp} shows the results of the MVC problem. The solutions generated by our methods achieve high coverage and the solution sizes are close to the optimal ones, especially for LP+{\namemodel}. For example, the solution generated by LP+{\namemodel} on Pubmed is optimal.
As expected, Table \ref{tab:misresultreal}  also shows that LP+{\namemodel} outperforms LP+{\namemodel$_{pt}$} on the MIS problem, and the solution sizes are close to the optimal ones. We attribute the high-quality solutions to the well-reduced search space that increases the purity of \textit{good nodes}.

\begin{table}[!htbp]
	\small
	\setlength\tabcolsep{4.5pt}
	\centering
	\begin{tabular}{l|lll}
		\toprule
		\textbf{Datasets} & LP & \makecell[l]{LP + \\ \namemodel$_{pt}$} & \makecell[l]{LP + \\ \namemodel} \\
		\midrule
		\textbf{OTC} & 1535 & 1509 (99.84) & 1538 (\textbf{100.00}) \\
		\textbf{Pubmed} & 3805 & 3789 (99.96) & 3806 (\textbf{100.00}) \\
		\textbf{Brightkite} & 21867 & 21141 (99.61) & 21918 (\textbf{99.98}) \\
		\textbf{Slashdot0811} & 24046 & 23903 (99.92) & 24098 (\textbf{99.98}) \\
		\textbf{Slashdot0922} & 25770 & 25698 (99.93) & 25928 (\textbf{99.99}) \\
		\textbf{Gowalla} & 84223 & 82874 (99.82) & 84471 (\textbf{99.99}) \\
		\bottomrule
	\end{tabular}
	\caption{MVC results (solution size with coverage of all edges in the parenthesis) of LP on real-world datasets. {\namemodel} always outperforms {\namemodel$_{pt}$} in terms of coverage and obtains the best solution on Pubmed.}
	\label{tab:mvcresultreallp}
\end{table}

\begin{table*}[!htbp]
	\small
	\centering
	\begin{tabular}{l|cccccc}
		\toprule
		\textbf{Method} & \textbf{OTC} & \textbf{Pubmed} & \textbf{Brightkite} & \textbf{Slashdot0811} & \textbf{Slashdot0922} & \textbf{Gowalla} \\
		\midrule
		LP & 4346 & 15912 & 36361 & 53314 & 56398 & 112363 \\
		LP+\namemodel$_{pt}$ & 4222 & 15637 & 32981 & 48445 & 50502 & 89790 \\
		LP+\namemodel & \textbf{4289} & \textbf{15896} & \textbf{35776} & \textbf{50903} & \textbf{53848} & \textbf{110569} \\
		\midrule
		GD & 4342 & 15912 & 36119 & 53314 & 56393 & 112214 \\
		GD+\namemodel$_{pt}$ & 4222 & 15637 & 32981 & 48445 & 50502 & 89790 \\
		GD+\namemodel & \textbf{4287} & \textbf{15896} & \textbf{35767} & \textbf{50903} & \textbf{53845} & \textbf{110462} \\
		\midrule
		LS & 4100 & 14501 & 34750 & 51323 & 54184 & 102349 \\
		LS+\namemodel$_{pt}$ & 4222 & 15637 & 32981 & 48445 & 50502 & 89790 \\
		LS+\namemodel & \textbf{4288} & \textbf{15873} & \textbf{35608} & \textbf{50897} & \textbf{53840} & \textbf{108606} \\
		\bottomrule
	\end{tabular}
	\caption{MIS results (solution size) on real-world datasets and the best results are marked in bold. Baseline+{\namemodel} outperforms Baseline+{\namemodel$_{pt}$} and even generates better solutions than the Baseline in some cases.}
	\label{tab:misresultreal}
\end{table*}

\begin{figure}[!htbp]
	\centering
	\begin{subfigure}{0.32\linewidth}
		\includegraphics[width=1.0\linewidth]{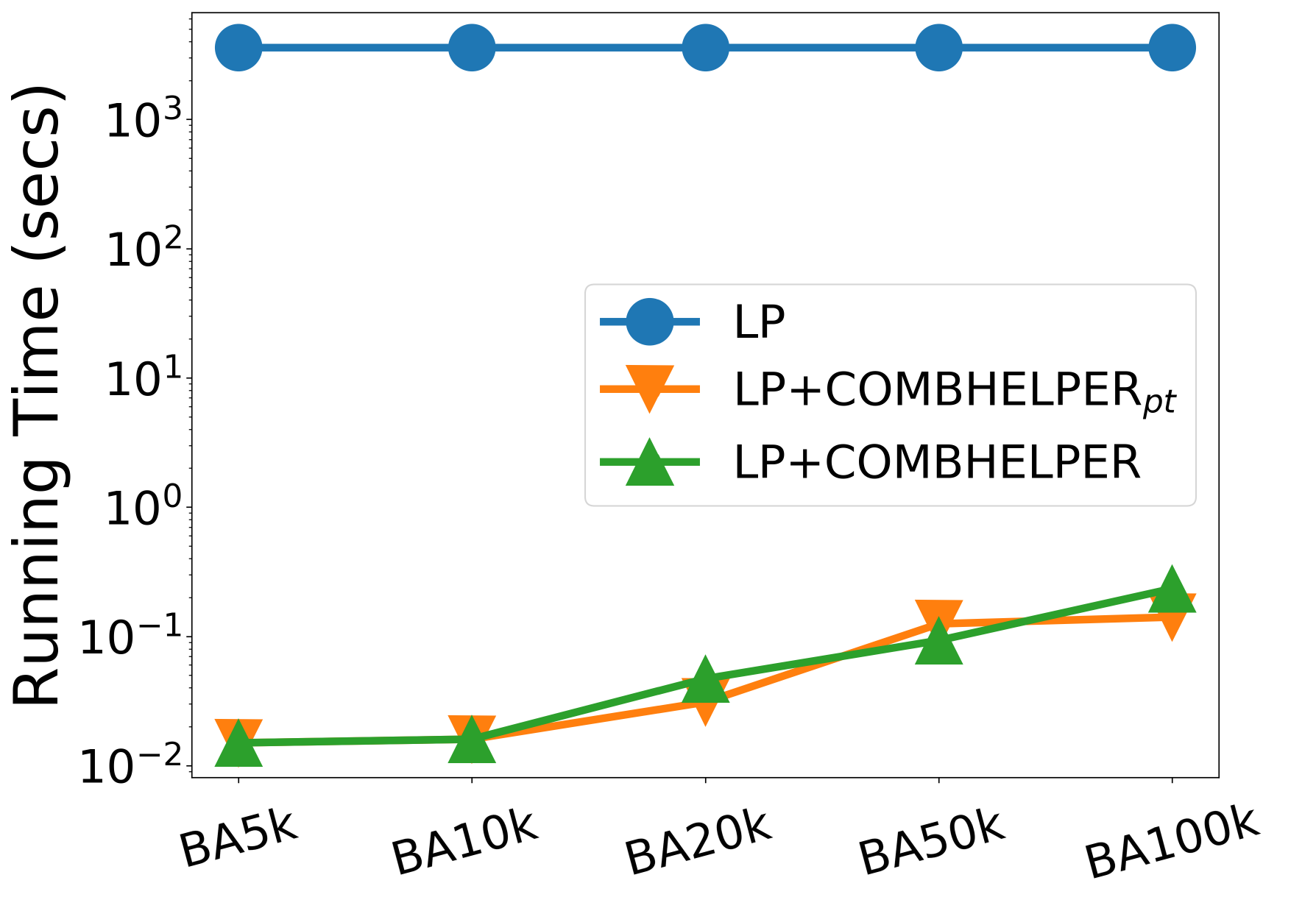}
		\caption{LP on MVC}
		\label{subfig:mvclpsynthetic}
	\end{subfigure}
	\centering
	\begin{subfigure}{0.32\linewidth}
		\includegraphics[width=1.0\linewidth]{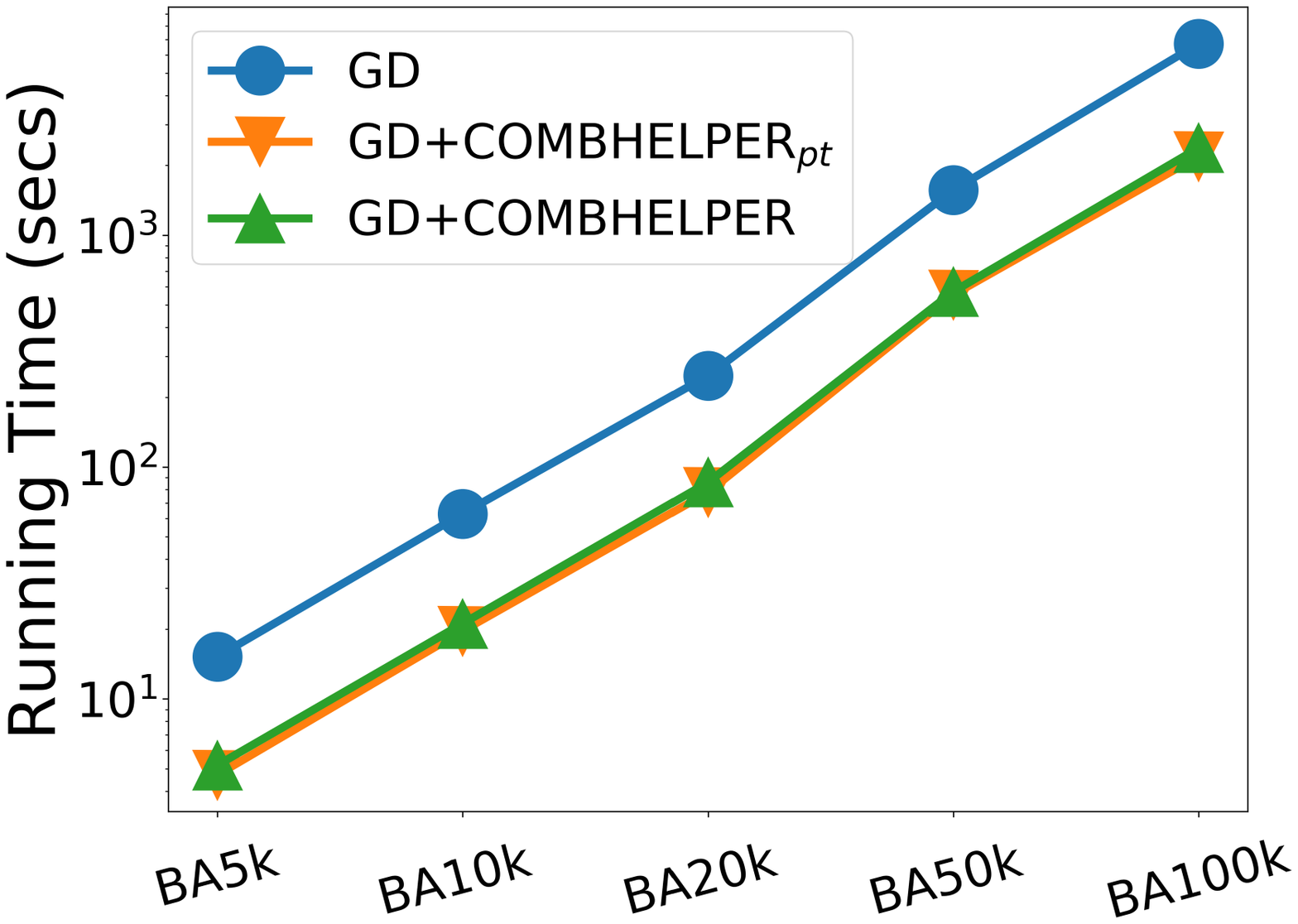}
		\caption{GD on MVC}
		\label{subfig:mvcgreedysynthetic}
	\end{subfigure}
	\centering
	\begin{subfigure}{0.32\linewidth}
		\includegraphics[width=1.0\linewidth]{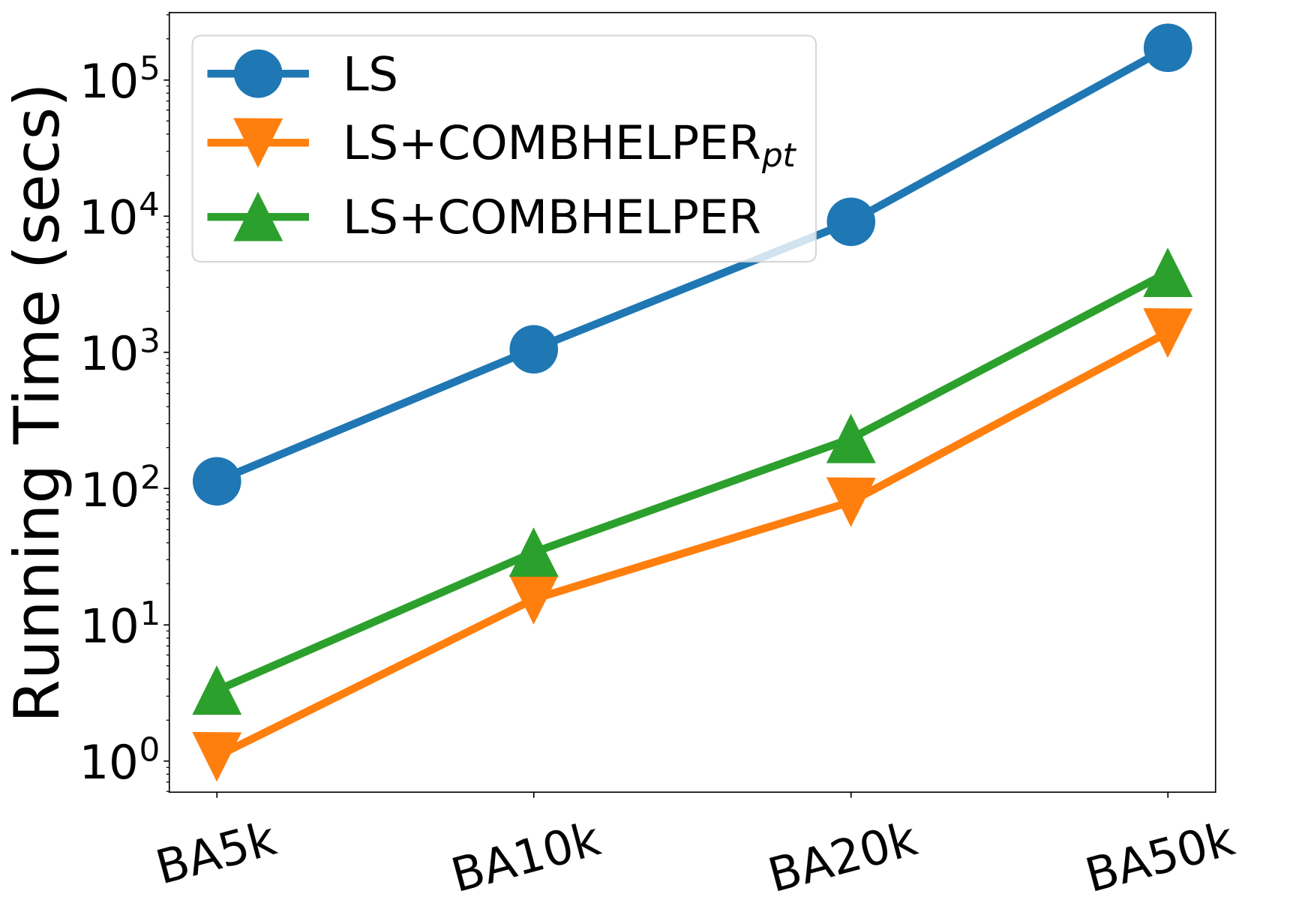}
		\caption{LS on MVC}
		\label{subfig:mvclssynthetic}
	\end{subfigure}
	
	\centering
	\begin{subfigure}{0.32\linewidth}
		\includegraphics[width=1.0\linewidth]{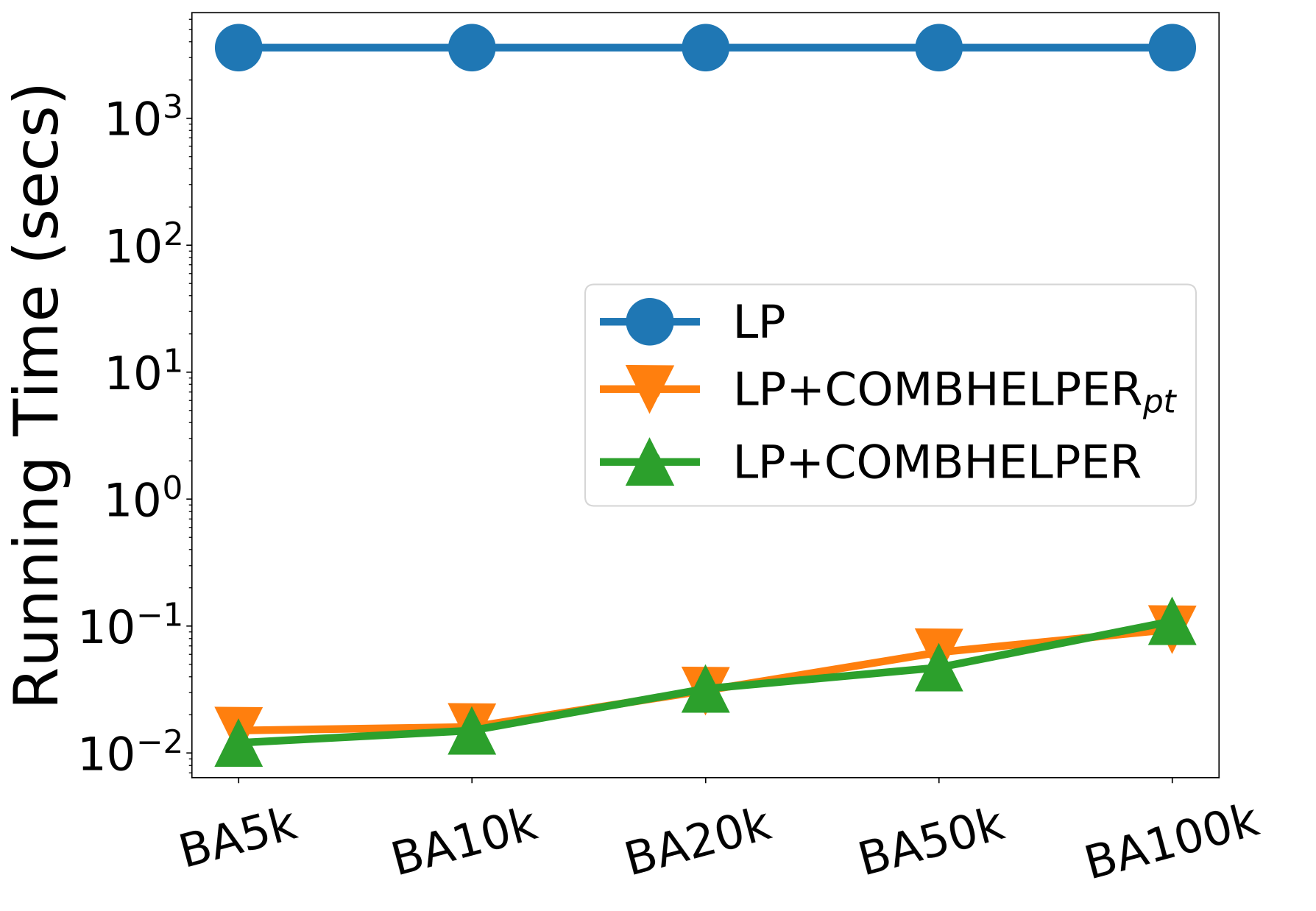}
		\caption{LP on MIS}
		\label{subfig:mislpsynthetic}
	\end{subfigure}
	\centering
	\begin{subfigure}{0.32\linewidth}
		\includegraphics[width=1.0\linewidth]{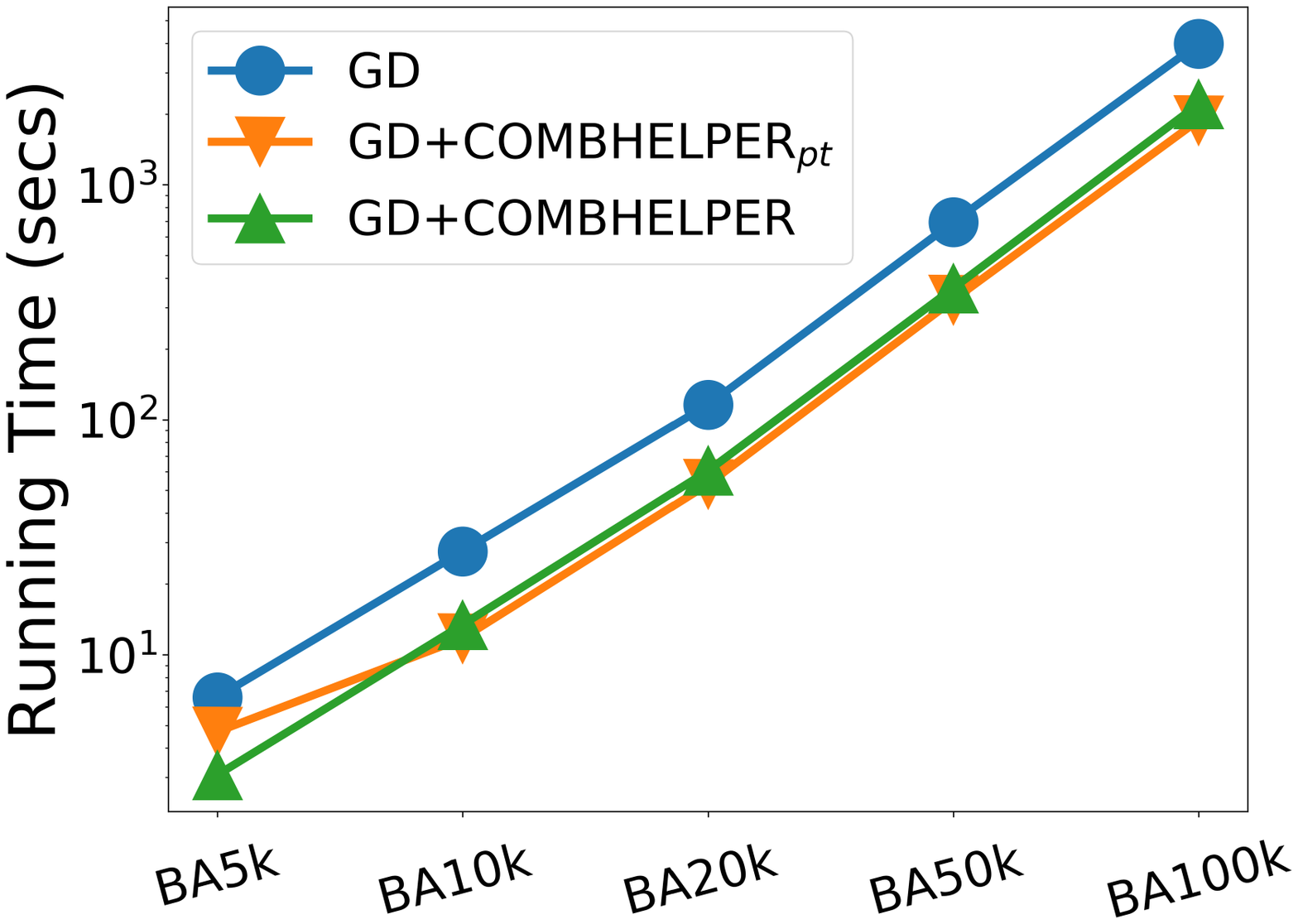}
		\caption{GD on MIS}
		\label{subfig:misgreedysynthetic}
	\end{subfigure}
	\centering
	\begin{subfigure}{0.32\linewidth}
		\includegraphics[width=1.0\linewidth]{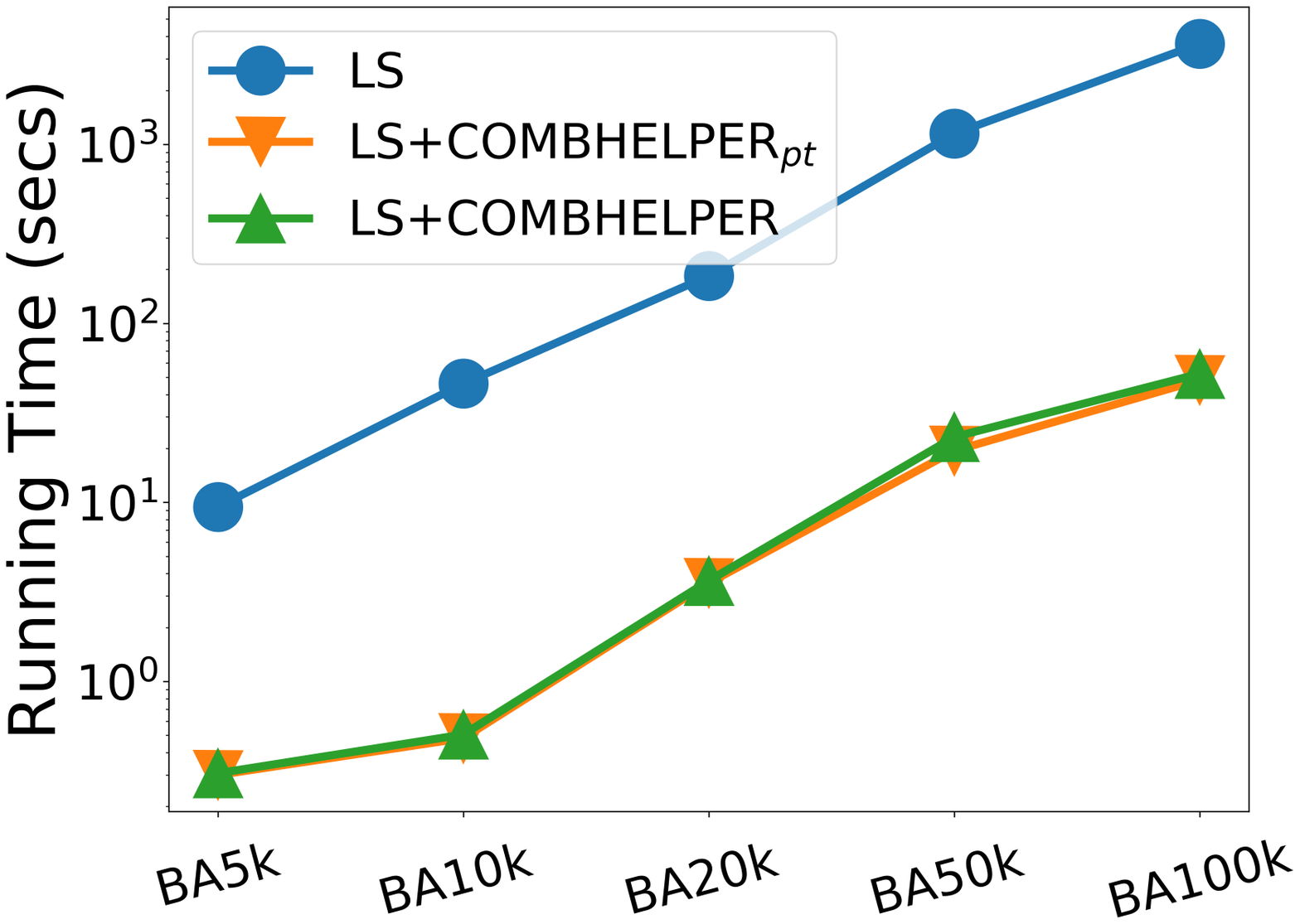}
		\caption{LS on MIS}
		\label{subfig:mislssynthetic}
	\end{subfigure}
	\caption{Running times (seconds) on synthetic datasets. Our method takes less time to generate solutions for both the MVC and MIS problems. Note that all the running times of general LP (blue line) in Figure \ref{subfig:mvclpsynthetic} and \ref{subfig:mislpsynthetic} reach the time limit of 1 hour. }
	\label{fig:runningtimesynthetic}
\end{figure}

\begin{figure}[!htbp]
	\centering
	\includegraphics[width=0.95\linewidth]{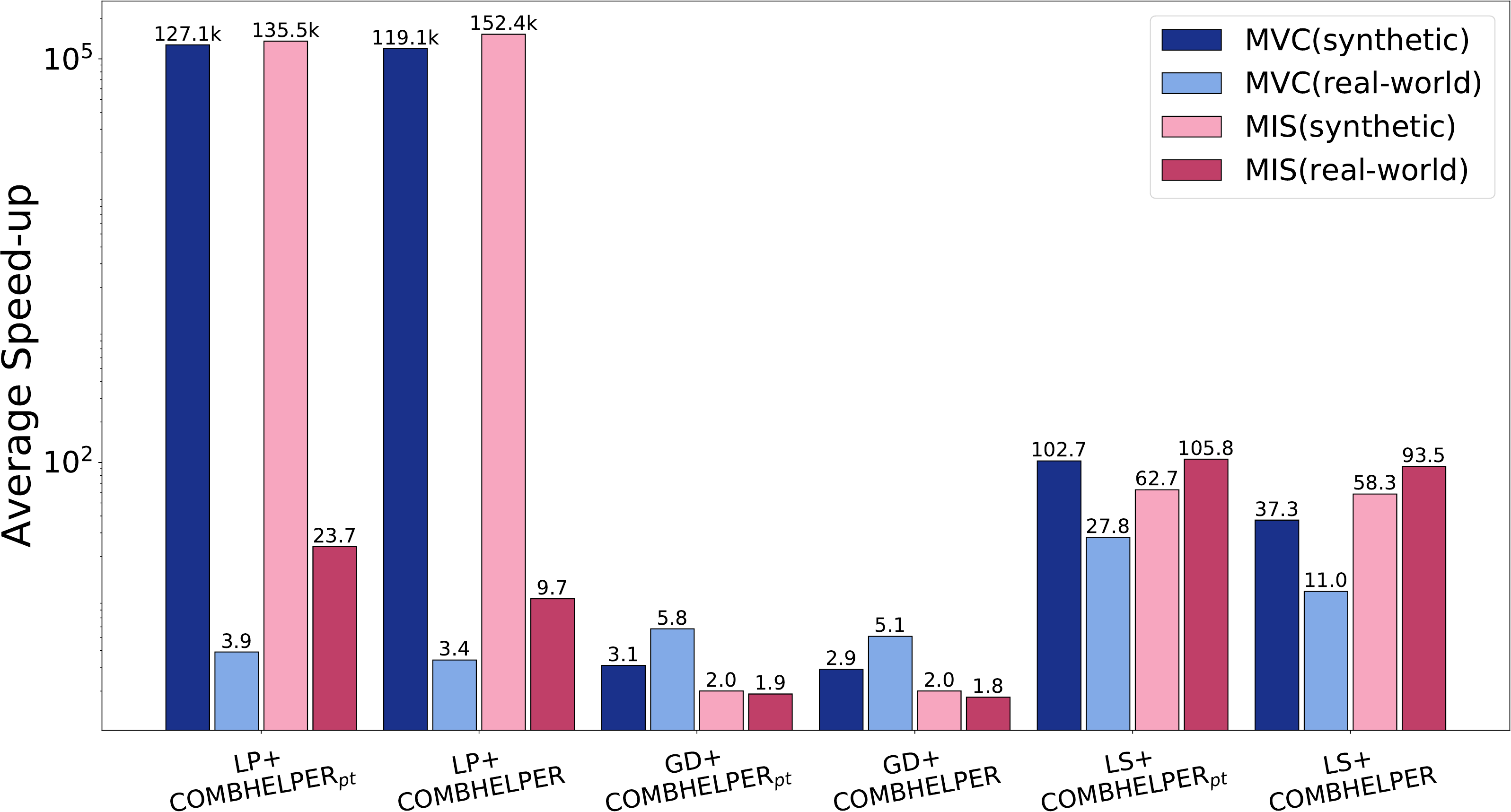}
	\caption{Average speed-up on both synthetic and real-world datasets. Baselines (LP, GD, and LS) with {\namemodel$_{pt}$} and {\namemodel} are at least 2 times faster than their original versions.}
	\label{fig:averagespeedup}
\end{figure}

\begin{figure}[!htbp]
	\centering
	\begin{subfigure}{0.32\linewidth}
		\includegraphics[width=1.0\linewidth]{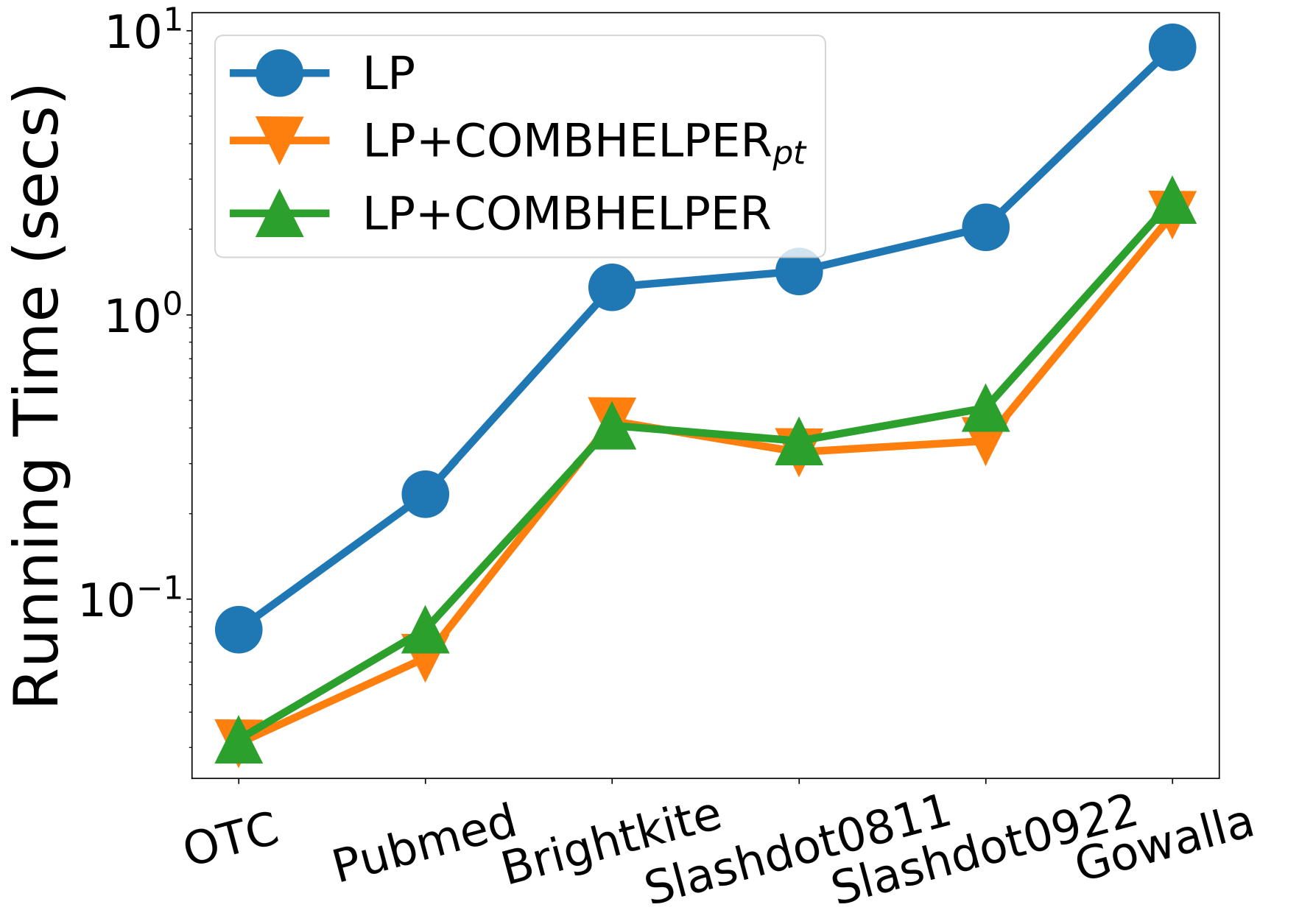}
		\caption{LP on MVC}
		\label{subfig:mvclpreal}
	\end{subfigure}
	\centering
	\begin{subfigure}{0.32\linewidth}
		\includegraphics[width=1.0\linewidth]{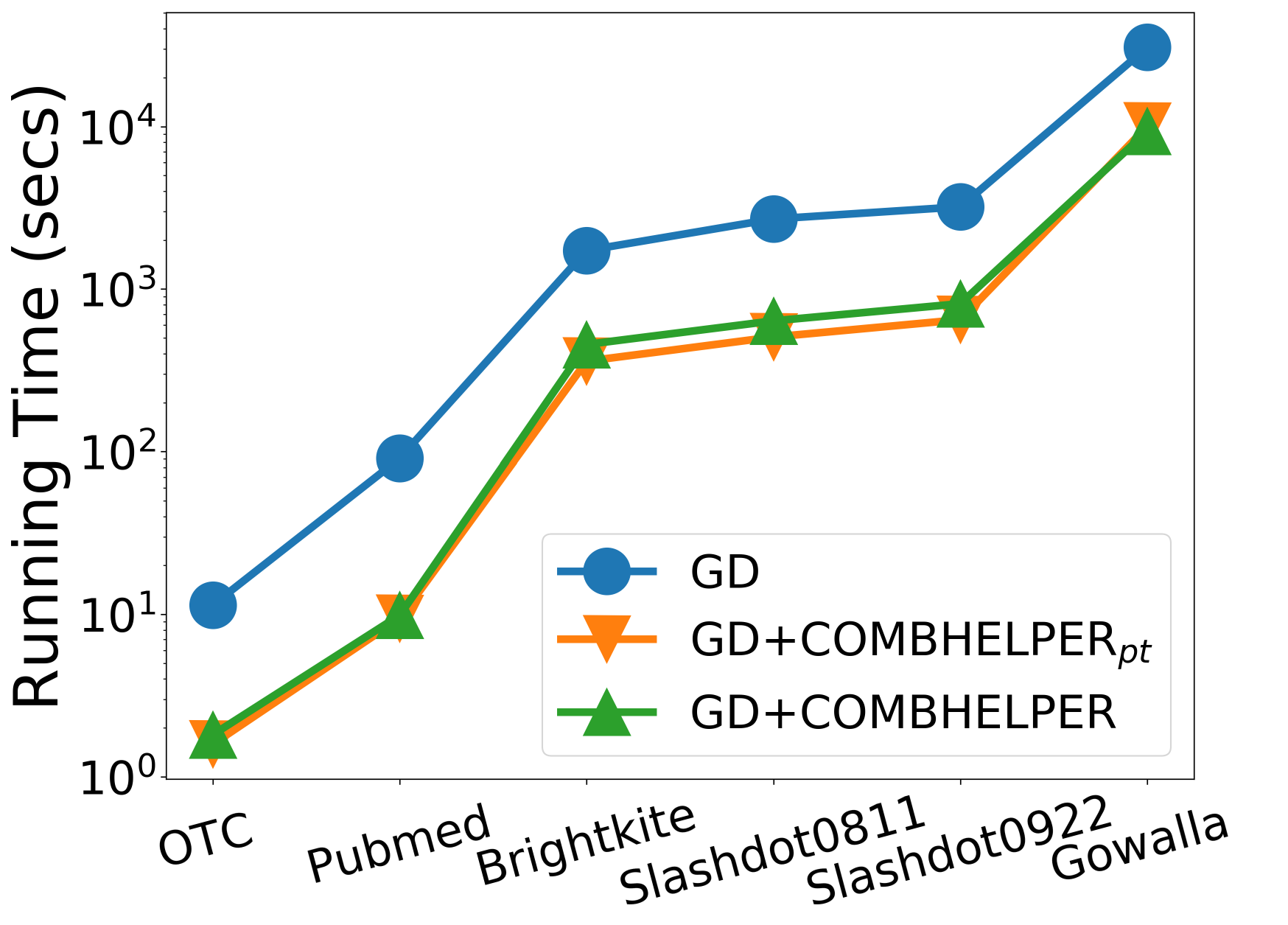}
		\caption{GD on MVC}
		\label{subfig:mvcgreedyreal}
	\end{subfigure}
	\centering
	\begin{subfigure}{0.32\linewidth}
		\includegraphics[width=1.0\linewidth]{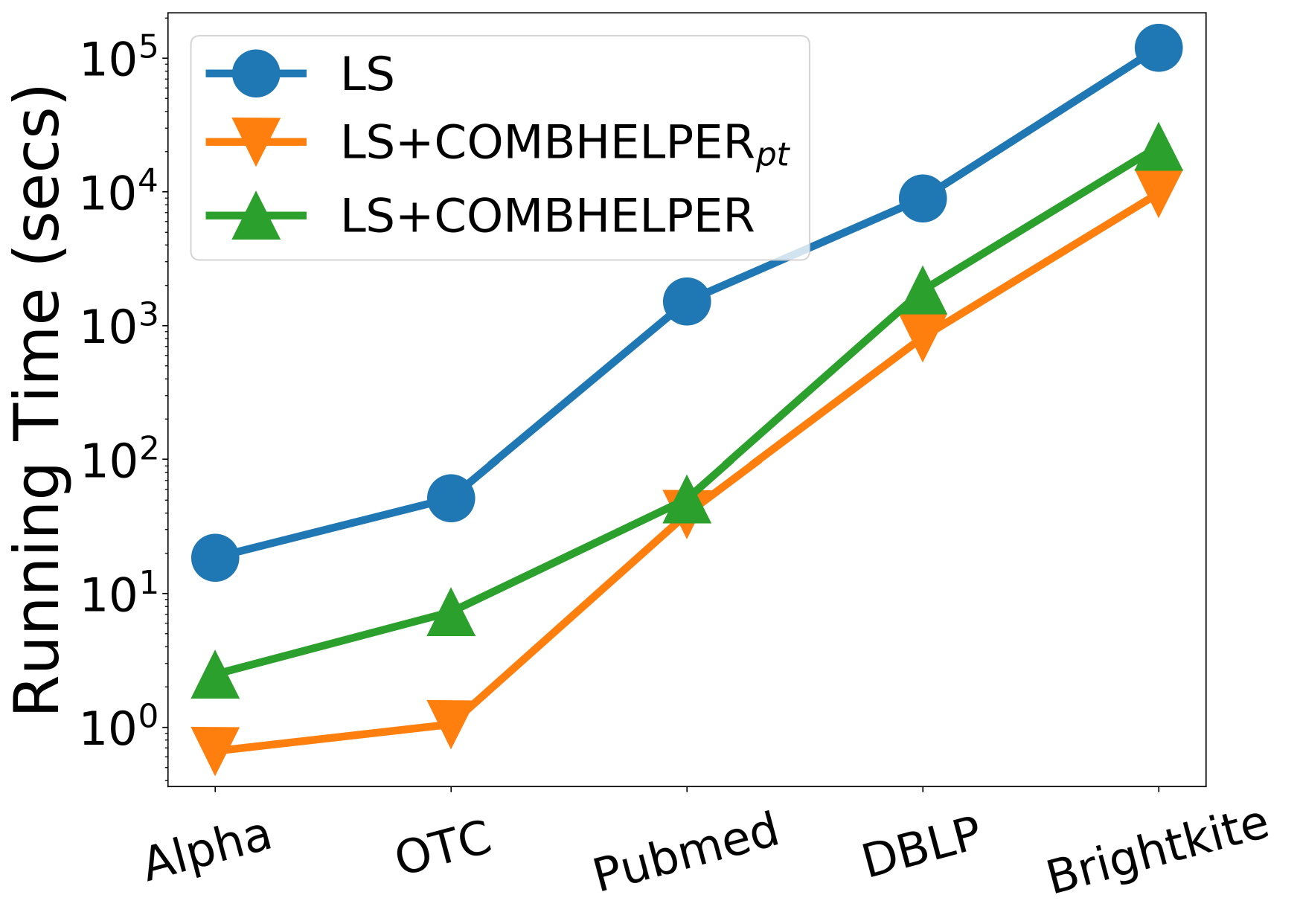}
		\caption{LS on MVC}
		\label{subfig:mvclsreal}
	\end{subfigure}
	
	\centering
	\begin{subfigure}{0.32\linewidth}
		\includegraphics[width=1.0\linewidth]{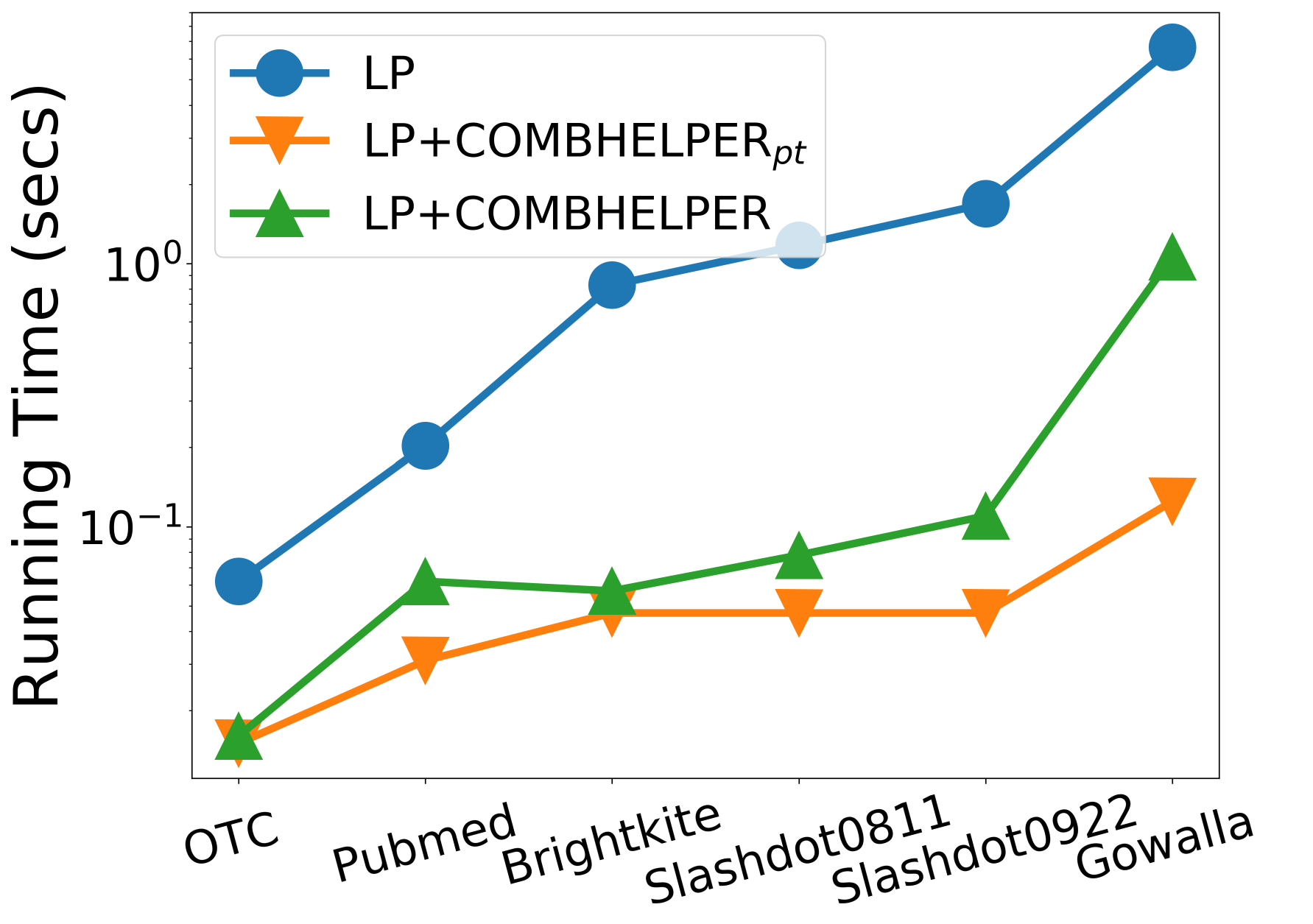}
		\caption{LP on MIS}
		\label{subfig:mislpreal}
	\end{subfigure}
	\centering
	\begin{subfigure}{0.32\linewidth}
		\includegraphics[width=1.0\linewidth]{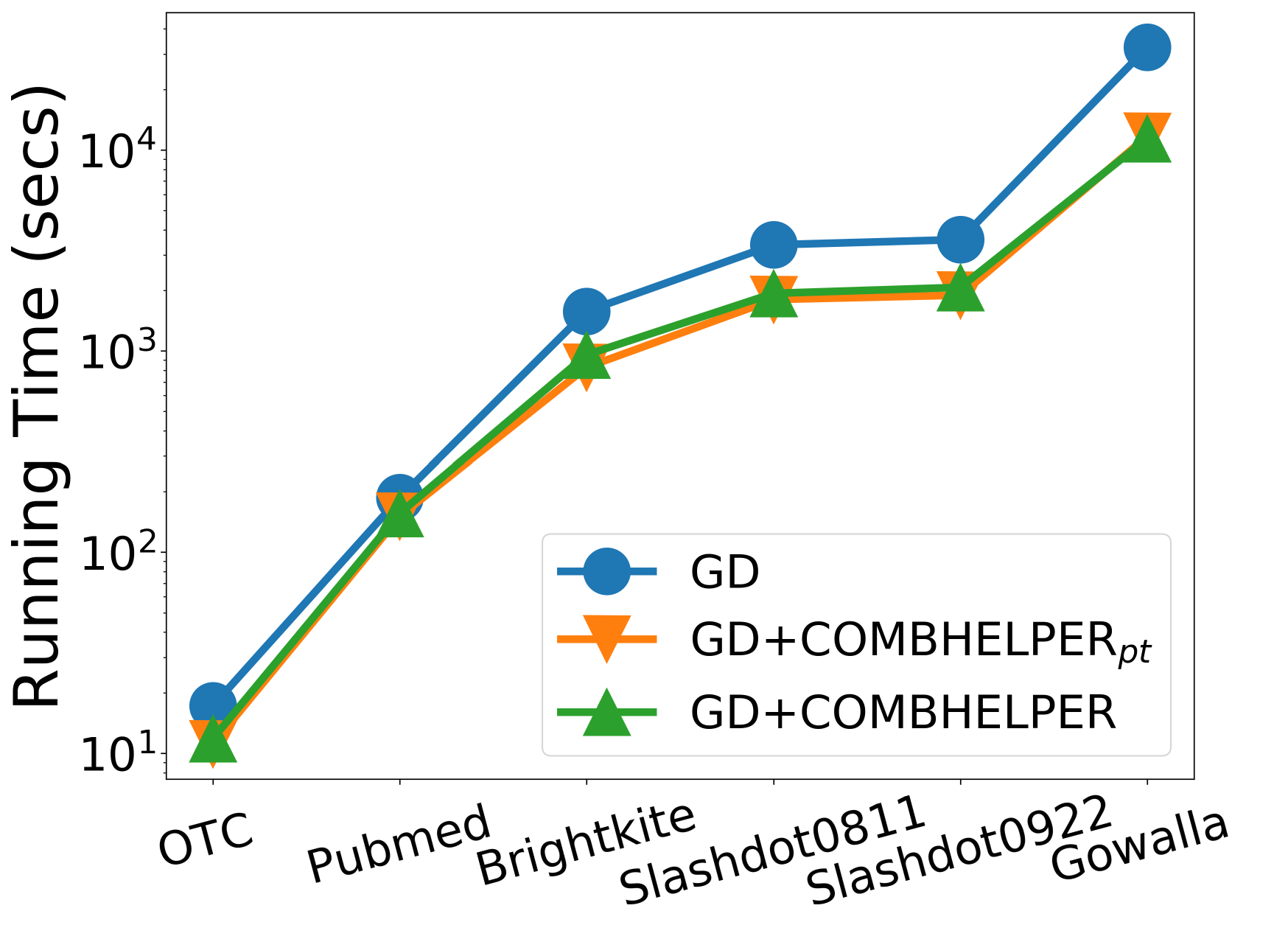}
		\caption{GD on MIS}
		\label{subfig:misgreedyreal}
	\end{subfigure}
	\centering
	\begin{subfigure}{0.32\linewidth}
		\includegraphics[width=1.0\linewidth]{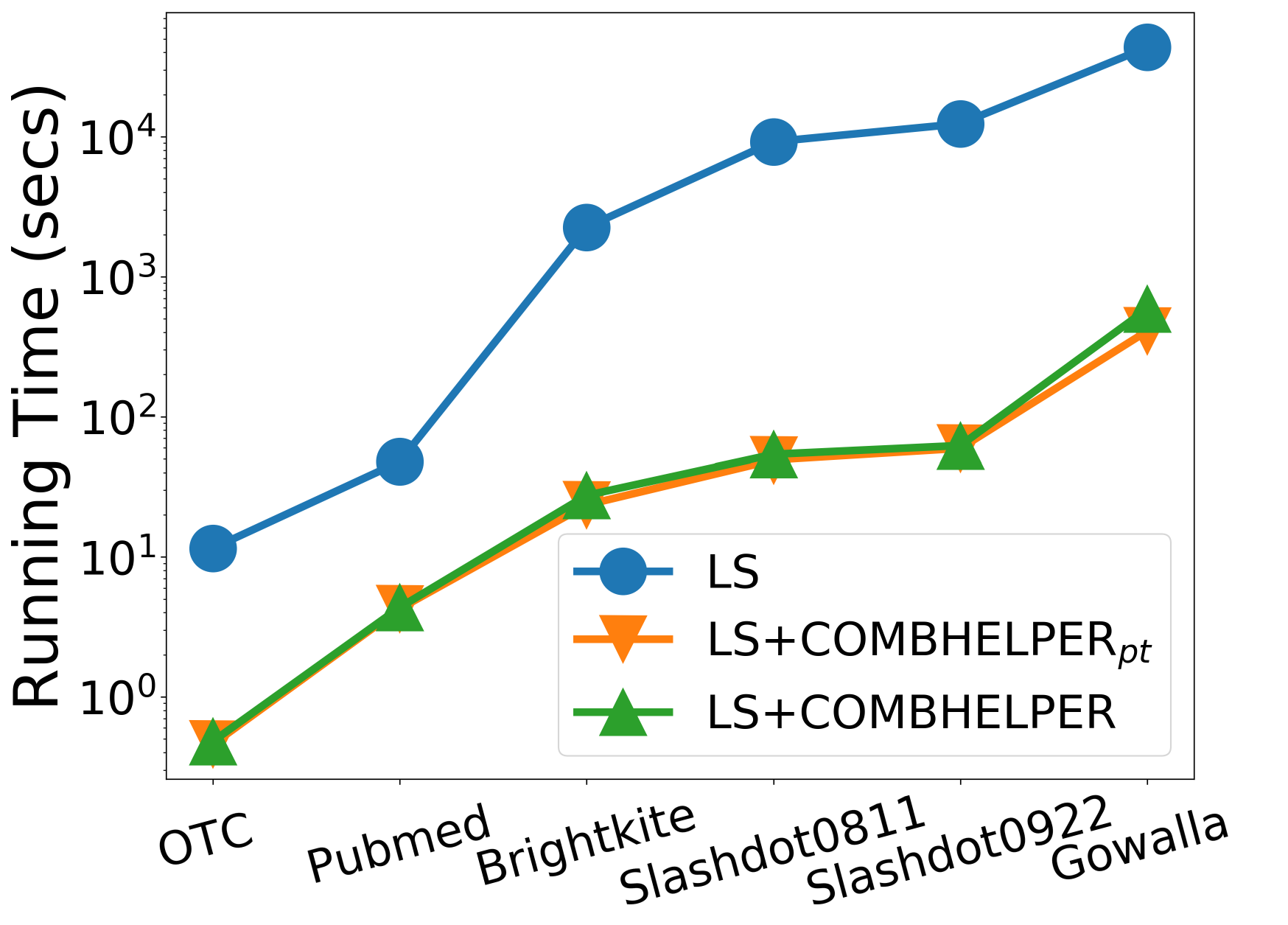}
		\caption{LS on MIS}
		\label{subfig:mislsreal}
	\end{subfigure}
	\caption{Running times (seconds) on real-world datasets. Our method generates solutions for both the MVC and MIS problems more efficiently.}
	\label{fig:runningtimereal}
\end{figure}

After showing the superior quality of our method, we now investigate its efficiency. Figure \ref{fig:runningtimesynthetic} and \ref{fig:runningtimereal} present that LP with {\namemodel$_{pt}$} and {\namemodel} take significantly lower time to generate solutions for both the MVC and MIS problems than the general LP. On the synthetic datasets, general LP can not solve both problems within 1 hour but {\namemodel$_{pt}$} and {\namemodel} generate solutions in less than 1 second. For real-world datasets, Figure \ref{fig:averagespeedup} shows that LP with {\namemodel$_{pt}$} and {\namemodel} are about 4 times faster than the general LP on the MVC problem and gain a speed-up range from 9.7 to 23.7 on the MIS problem. Our method is more efficient since unlikely nodes are dropped, reducing the search space size of LP.

\begin{table}[!htbp]
	\small
	\setlength\tabcolsep{4.5pt}
	\centering
	\begin{tabular}{l|lll}
		\toprule
		\textbf{Datasets} & GD & \makecell[l]{GD + \\ \namemodel$_{pt}$} & \makecell[l]{GD + \\ \namemodel} \\
		\midrule
		\textbf{OTC} &1555 &1519 (99.84)&1551 (\textbf{100.00}) \\
		\textbf{Pubmed} &3866&3811 (99.96)&3846 (\textbf{100.00}) \\
		\textbf{Brightkite} &22159&21285 (99.61)&22090 (\textbf{99.98}) \\
		\textbf{Slashdot0811} & 24347 & 23995 (99.92) & 24339 (\textbf{99.98}) \\
		\textbf{Slashdot0922} & 26054 & 25828 (99.93) & 26182 (\textbf{99.99}) \\
		\textbf{Gowalla} & 85450 & 83694 (99.82) & 85343 (\textbf{99.99}) \\
		\bottomrule
	\end{tabular}
	\caption{MVC results (solution size with coverage of all edges in the parenthesis) of GD on real-world datasets. {\namemodel} always shows the best performance on GD.}
	\label{tab:mvcresultrealgd}
\end{table}

\subsubsection{Performance on GD.}

Greedy (GD) has been a popular algorithm for many graph CO problems. Similar to the case in LP, we demonstrate that our method improves GD. Table \ref{tab:misresultreal} and \ref{tab:mvcresultrealgd} present a similar conclusion that GD with both {\namemodel$_{pt}$} and {\namemodel} still generate high-quality solutions for both the MVC and MIS problems. In specific, GD+{\namemodel} obtains better MVC solutions (cover all the edges with fewer nodes) than the general GD on OTC and Pubmed, which means {\namemodel} obtains a better search space for solution generation.

Figure \ref{fig:runningtimesynthetic} and \ref{fig:runningtimereal} present the running times of GD and our methods. It is clear that both {\namemodel$_{pt}$} and {\namemodel} accelerate the process of solution generation in all the settings. Figure \ref{fig:averagespeedup} also indicates that GD with our method is about 2 to 6 times faster than the general GD, owing to the search space pruning procedure in {\namemodel}. We notice that the speed-up of GD is not so obvious as in LP and LS due to the limitation of GD itself, i.e., the efficiency of GD relies on the nature of CO problems (see the supplementary for further explanation).

\begin{table}[!htbp]
	\small
	\centering
	\begin{tabular}{l|lll}
		\toprule
		\textbf{Datasets} & LS & \makecell[l]{LS + \\ \namemodel$_{pt}$} & \makecell[l]{LS + \\ \namemodel} \\
		\midrule
		\textbf{Alpha} & 1097 & 1050 (99.84) & 1072 (\textbf{100.00}) \\
		\textbf{OTC} & 1564 & 1517 (99.84) & 1552 (\textbf{100.00}) \\
		\textbf{Pubmed} & 3896 & 3799 (99.96) & 3819 (\textbf{100.00}) \\
		\textbf{DBLP} & 8323 & 7818 (99.03) & 8315 (\textbf{99.99}) \\
		\textbf{Brightkite} & 22622 & 21417 (99.96) & 22187 (\textbf{99.99}) \\
		\bottomrule
	\end{tabular}
	\caption{MVC results of LS on real-world datasets. Solutions generated by LS+{\namemodel} are the best.}
	\label{tab:mvcresultrealls}
\end{table}

\subsubsection{Performance on LS.}

Similar observations can be seen from another popular CO algorithm called local search (LS). Since LS costs about 32 and 48 hours to generate MVC solutions on Brightkite and BA50k respectively, we do not test {\namemodel} on Slashdot0811, Slashdot0922, Gowalla, and BA100k for the MVC problem. Instead, we use another two small datasets Alpha and DBLP. 

Table \ref{tab:misresultreal} and \ref{tab:mvcresultrealls} demonstrate that LS+{\namemodel} generate the best solutions in most cases. Specifically, for the MVC problem, the solution sizes of LS+{\namemodel} on Alpha, OTC, and Pubmed are smaller than those of LS. For the MIS problem, the solution sizes of LS+{\namemodel} are larger than those of LS on all the datasets except Slashdot0811 and Slashdot0922.

From Figure \ref{fig:runningtimesynthetic} and \ref{fig:runningtimereal}, it is easy to see that LS with {\namemodel$_{pt}$} and {\namemodel} obtain the solutions more efficiently. Figure \ref{fig:averagespeedup} also shows that LS with {\namemodel$_{pt}$} and {\namemodel} gain a speed-up range from 11.0 to 102.7 on the MVC problem and are about 80 times faster than the general LS on the MIS problem. The major reason for such improvement in efficiency is that {\namemodel} accelerates both phases in LS: (1) initial solution generation and (2) neighborhood solution exploration.

\subsection{Ablation Study}

\subsubsection{Efficiency of KD.}

\begin{figure}[!htbp]
	\centering
	\begin{subfigure}{0.32\linewidth}
		\includegraphics[width=1.0\linewidth]{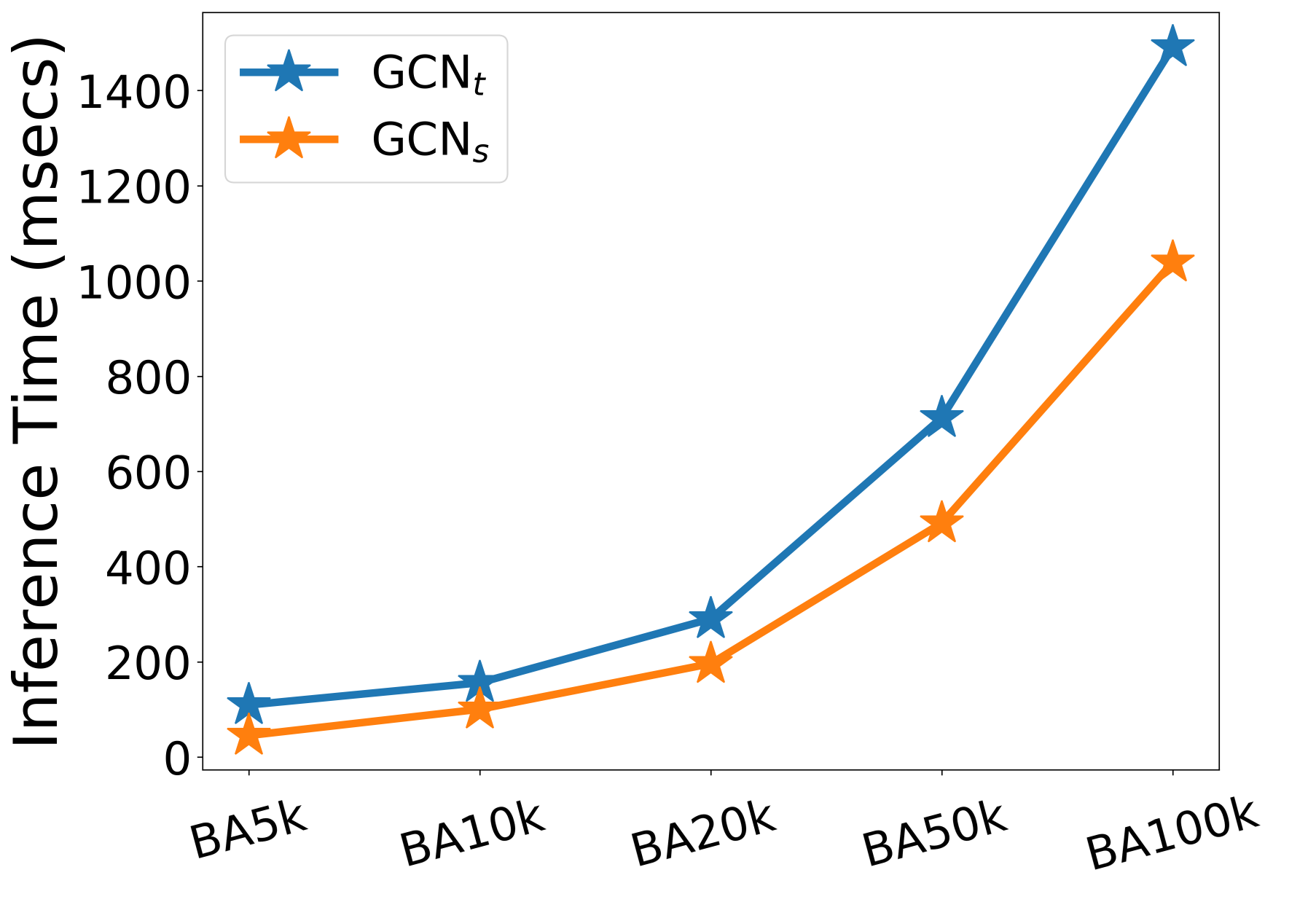}
		\caption{MVC}
		\label{subfig:inferencemvcsynthetic}
	\end{subfigure}
	\centering
	\begin{subfigure}{0.32\linewidth}
		\includegraphics[width=1.0\linewidth]{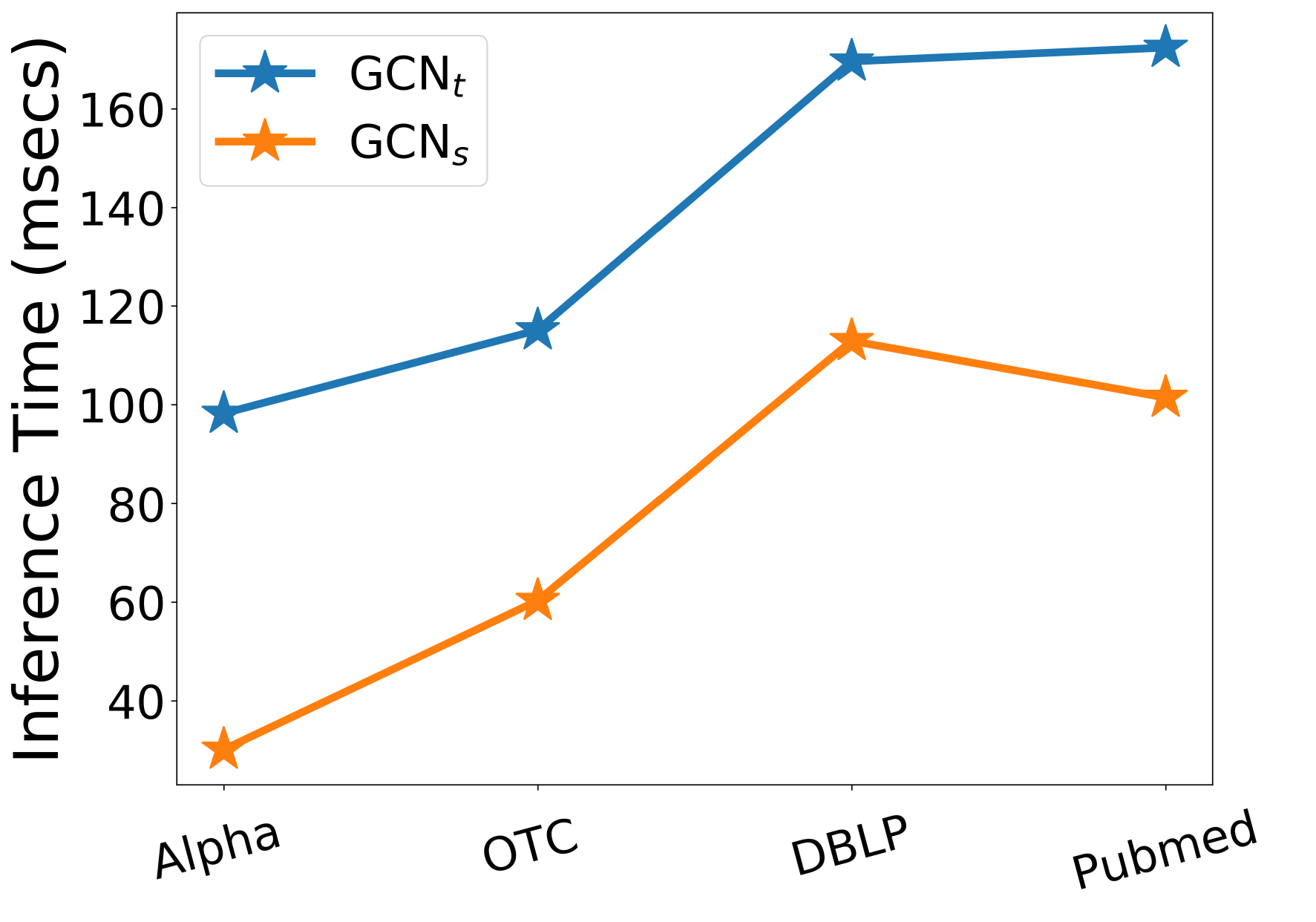}
		\caption{MVC}
		\label{subfig:inferencemvcreal1}
	\end{subfigure}
	\centering
	\begin{subfigure}{0.32\linewidth}
		\includegraphics[width=1.0\linewidth]{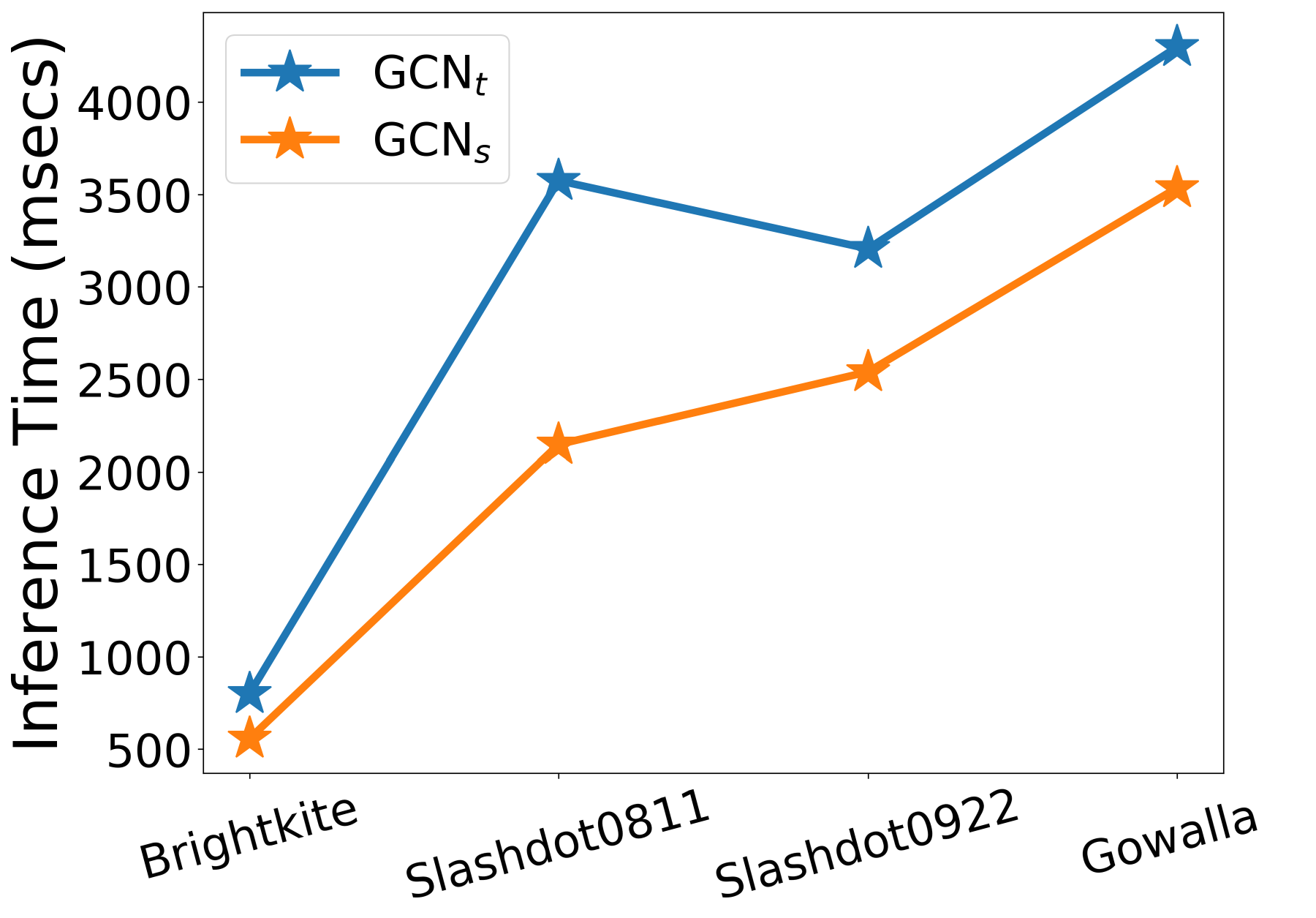}
		\caption{MVC}
		\label{subfig:inferencemvcreal2}
	\end{subfigure}
	
	\centering
	\begin{subfigure}{0.32\linewidth}
		\includegraphics[width=1.0\linewidth]{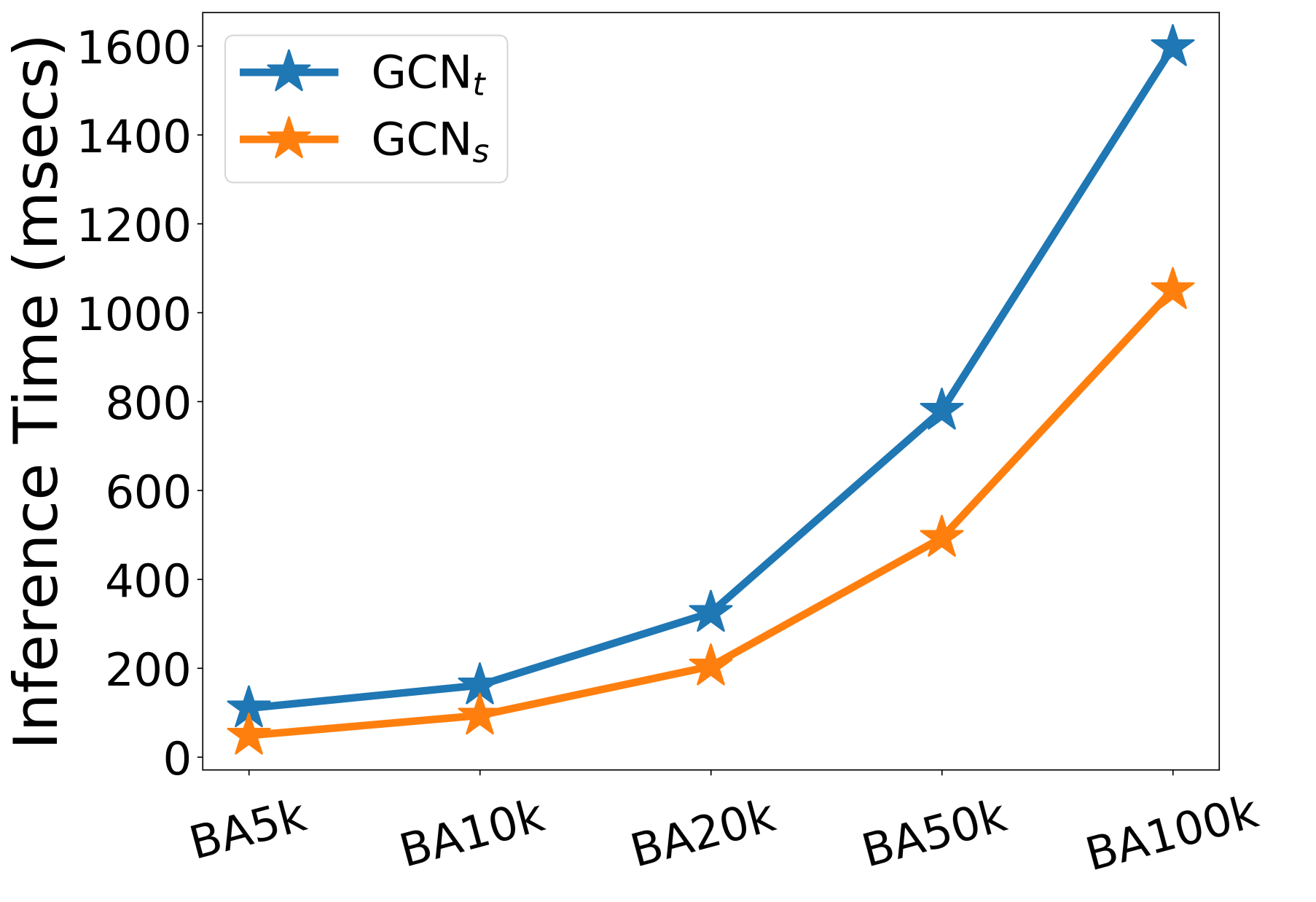}
		\caption{MIS}
		\label{subfig:inferencemissynthetic}
	\end{subfigure}
	\centering
	\begin{subfigure}{0.32\linewidth}
		\includegraphics[width=1.0\linewidth]{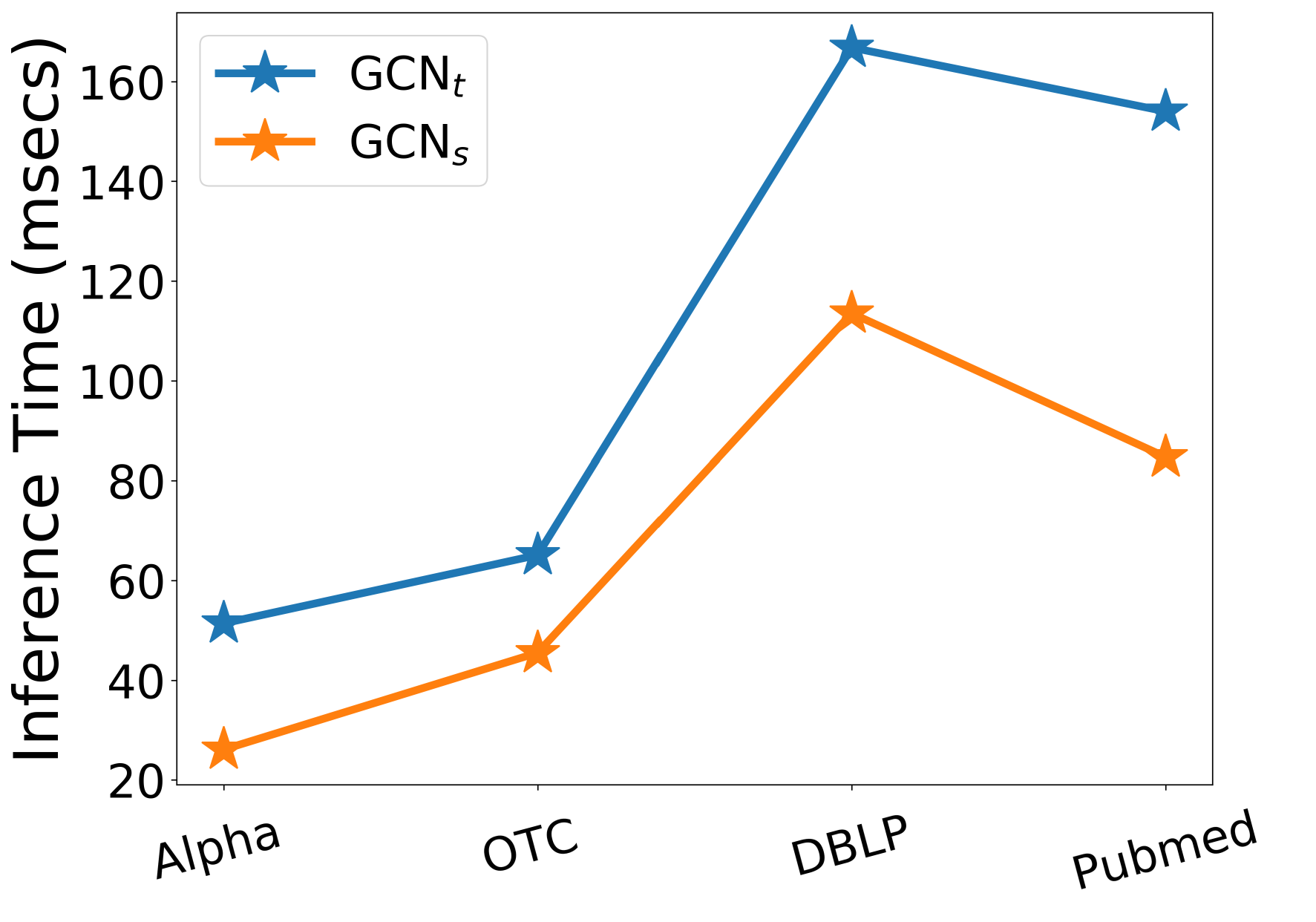}
		\caption{MIS}
		\label{subfig:inferencemisreal1}
	\end{subfigure}
	\centering
	\begin{subfigure}{0.32\linewidth}
		\includegraphics[width=1.0\linewidth]{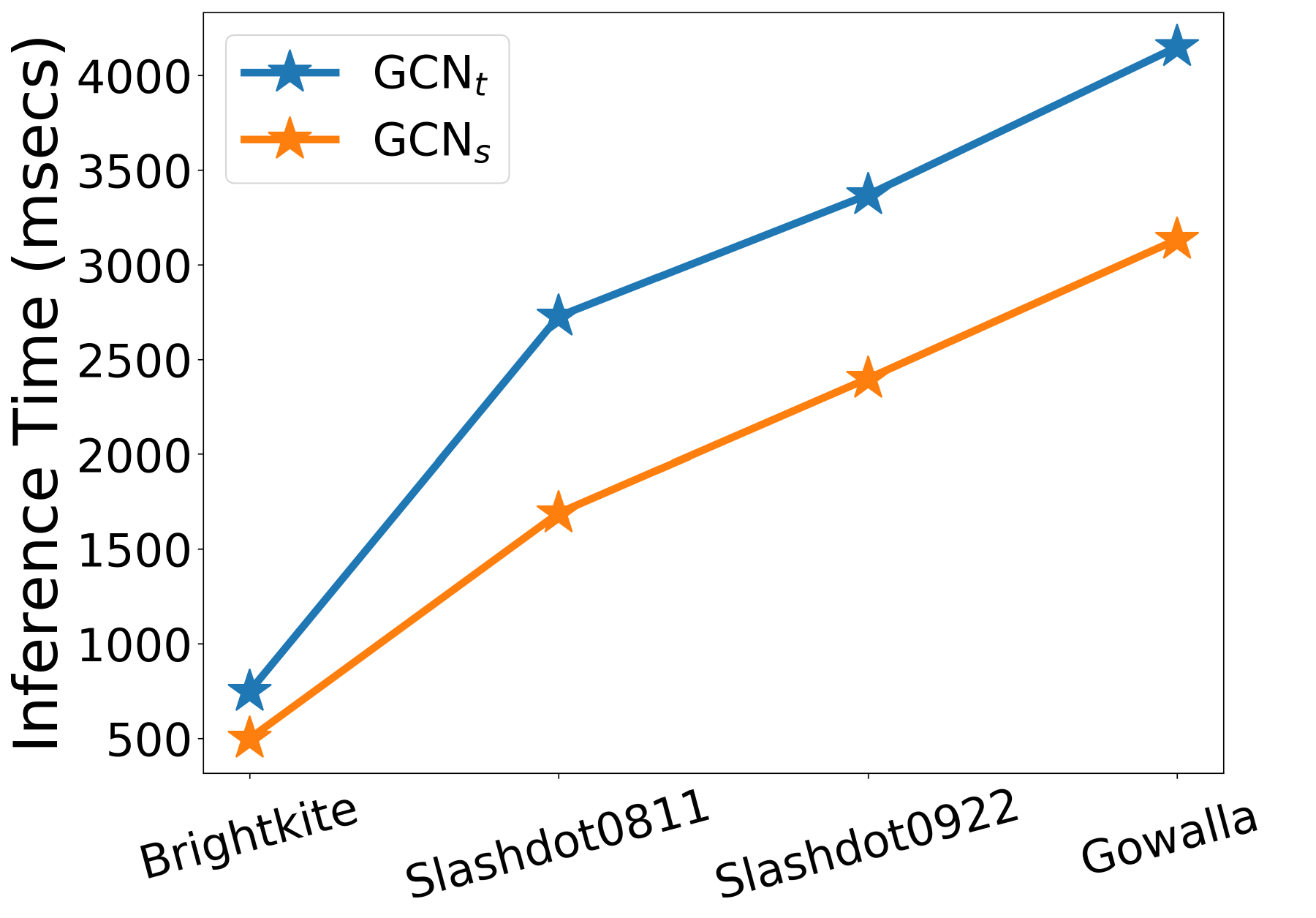}
		\caption{MIS}
		\label{subfig:inferencemisreal2}
	\end{subfigure}
	\caption{Inference time (milliseconds) on synthetic and real-world datasets. GCN$_{s}$ costs less time than GCN$_{t}$ in all the cases.}
	\label{fig:inferencetime}
\end{figure}

In order to verify the efficiency of KD, we compare the inference time (time of predicting \textit{good nodes}) of the teacher model (GCN$_{t}$) with that of the student model (GCN$_{s}$). From Figure \ref{fig:inferencetime}, it is obvious that GCN$_{s}$ costs less time than GCN$_{t}$ in all the sittings since the model compressed via KD has fewer parameters and a simpler structure, which predicts the \textit{good nodes} and reduce the search space more efficiently.

\subsubsection{Efficacy of Problem-Specific Boosting.}

In our case, if we can predict the nodes in the optimal solution as more as possible, we can prune the search space more precisely. So we use a widely used binary classification metric called \textit{recall} to evaluate the performance of GCN. It is formulated as $recall=\frac{TP}{TP+FN}$, where $TP$ means \textbf{T}rue \textbf{P}ositive and $FN$ means \textbf{F}alse \textbf{N}egative.

\begin{figure}[!htbp]
	\centering
	\begin{subfigure}{0.32\linewidth}
		\includegraphics[width=1.0\linewidth]{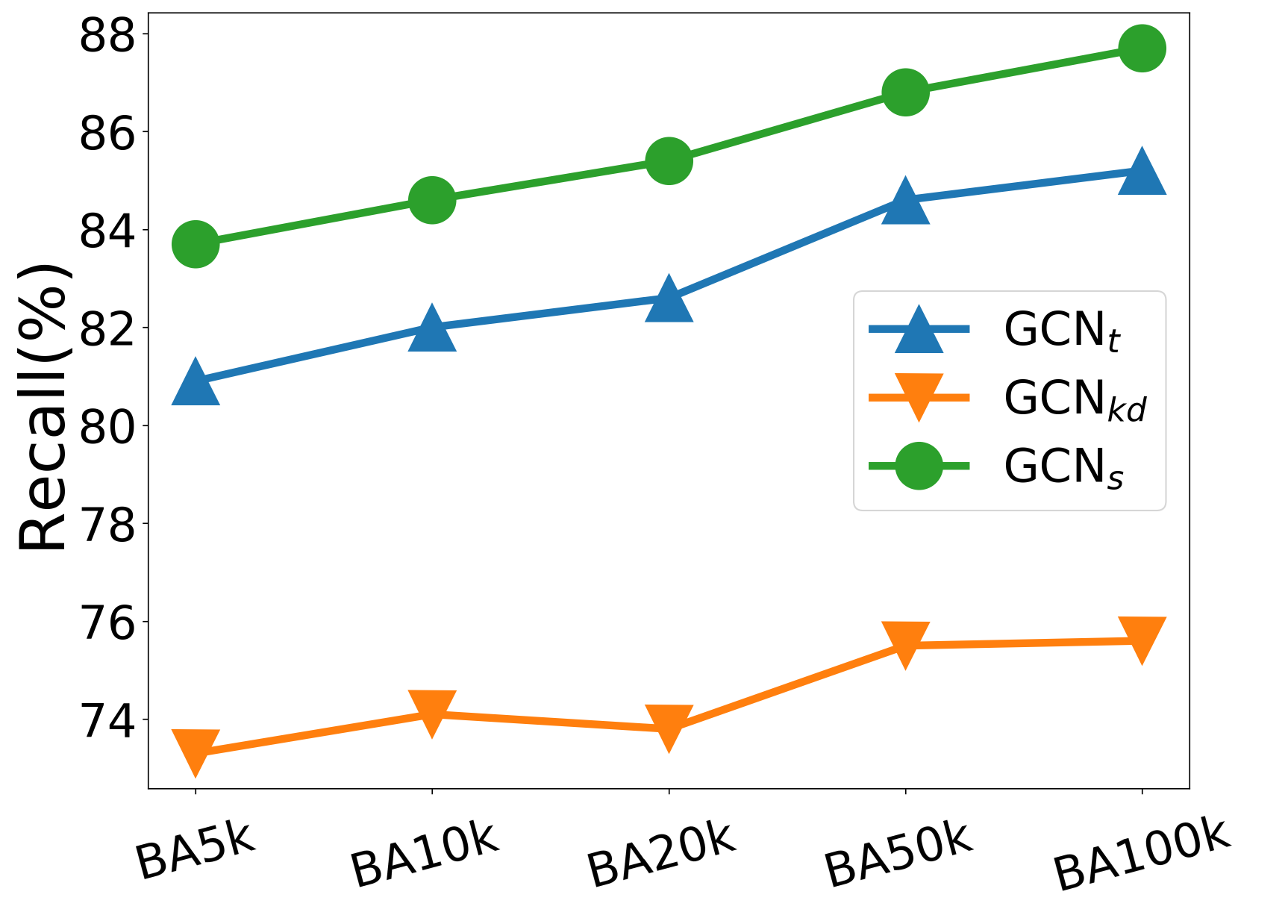}
		\caption{MVC}
		\label{subfig:recallmvcsynthetic}
	\end{subfigure}
	\centering
	\begin{subfigure}{0.32\linewidth}
		\includegraphics[width=1.0\linewidth]{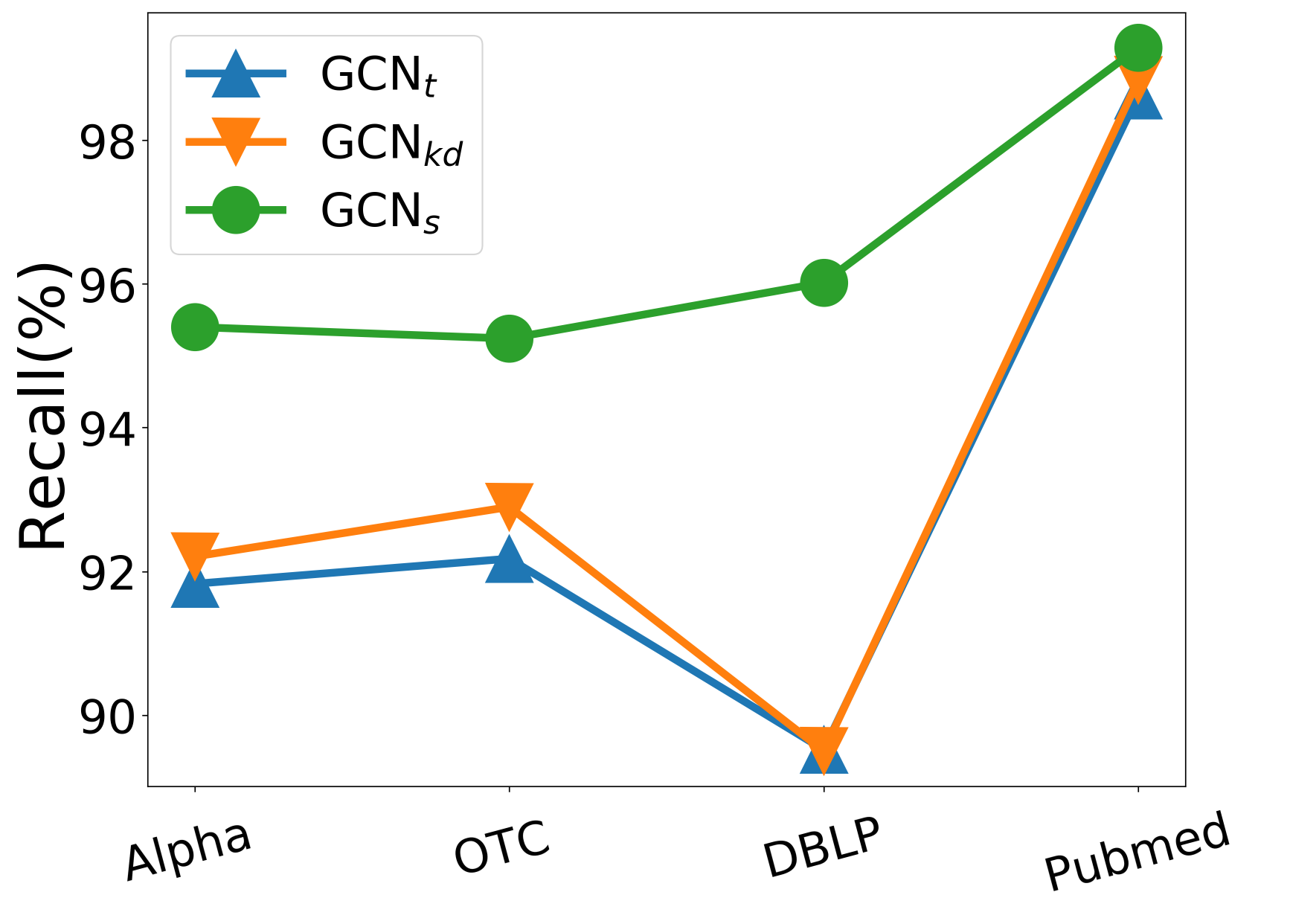}
		\caption{MVC}
		\label{subfig:recallmvcreal1}
	\end{subfigure}
	\centering
	\begin{subfigure}{0.32\linewidth}
		\includegraphics[width=1.0\linewidth]{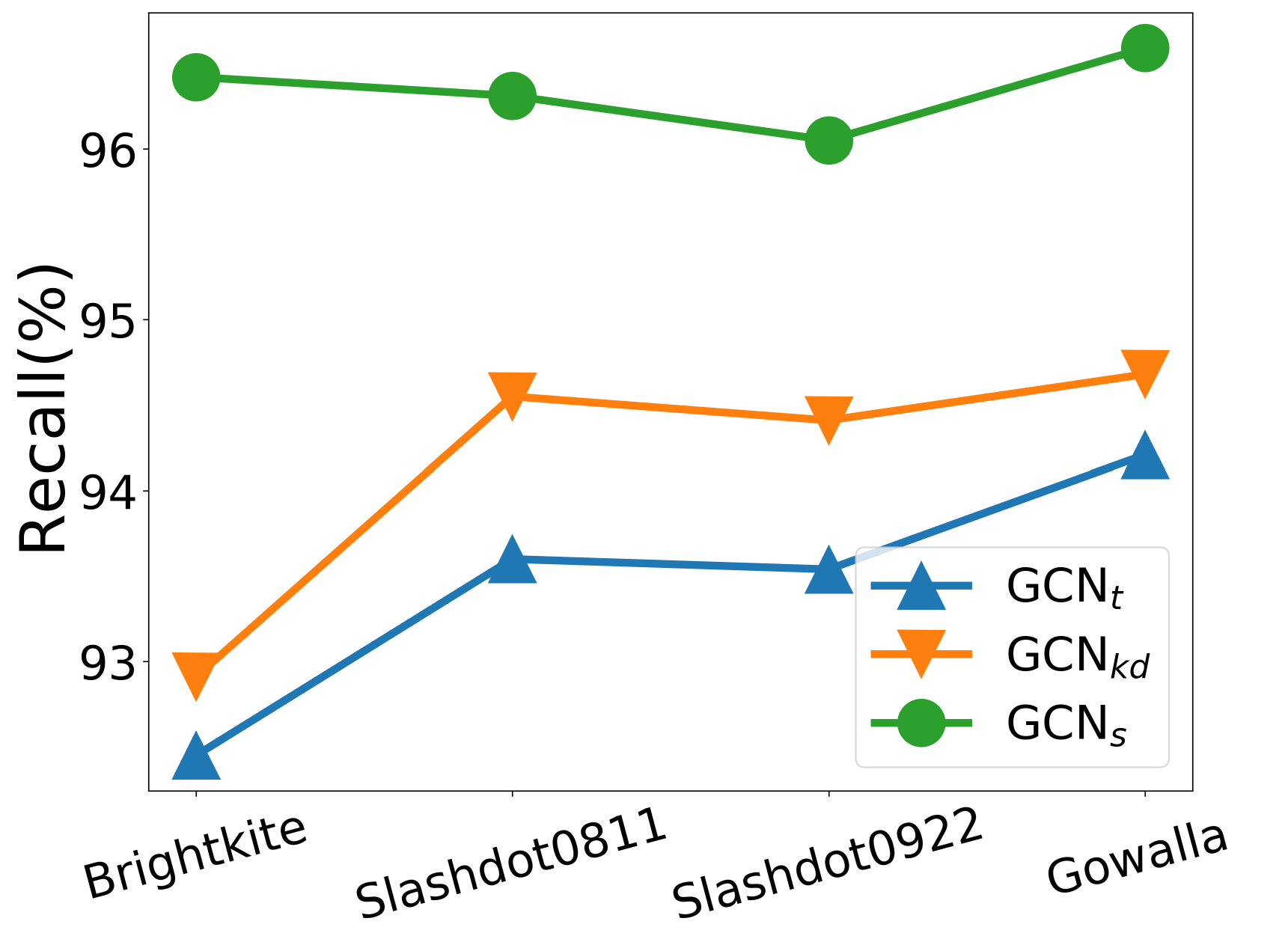}
		\caption{MVC}
		\label{subfig:recallmvcreal2}
	\end{subfigure}
	
	\begin{subfigure}{0.32\linewidth}
		\includegraphics[width=1.0\linewidth]{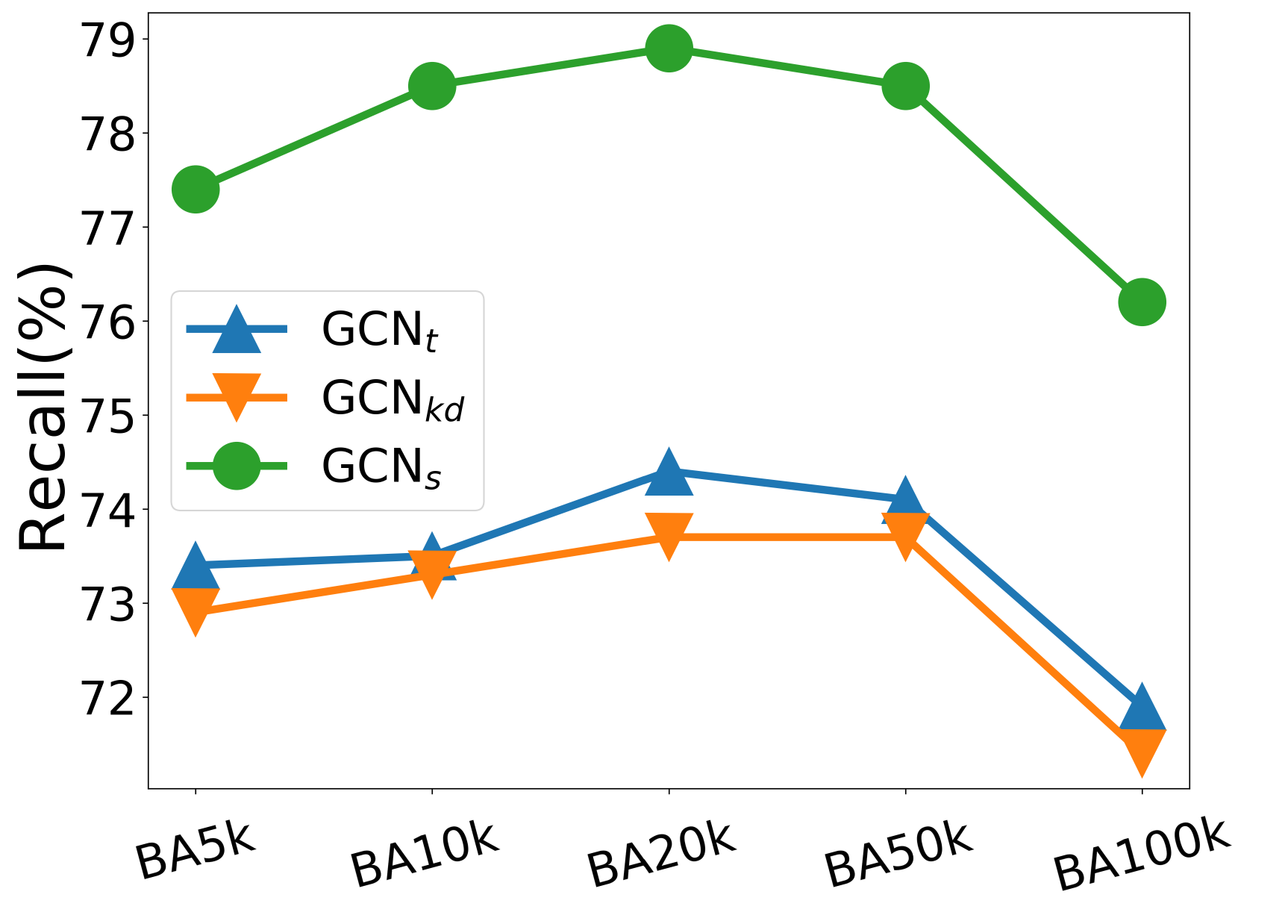}
		\caption{MIS}
		\label{subfig:recallmissynthetic}
	\end{subfigure}
	\centering
	\begin{subfigure}{0.32\linewidth}
		\includegraphics[width=1.0\linewidth]{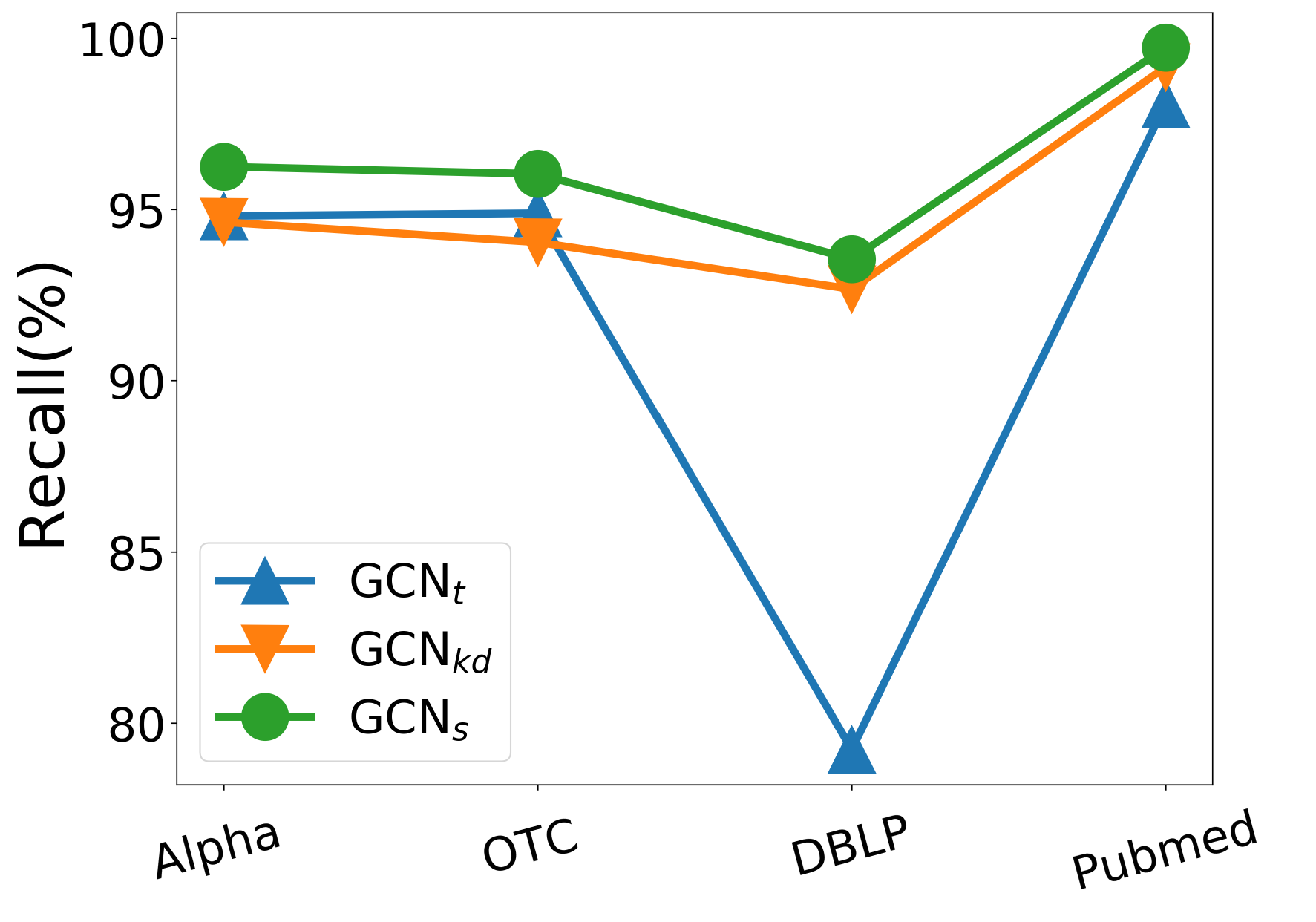}
		\caption{MIS}
		\label{subfig:recallmisreal1}
	\end{subfigure}
	\centering
	\begin{subfigure}{0.32\linewidth}
		\includegraphics[width=1.0\linewidth]{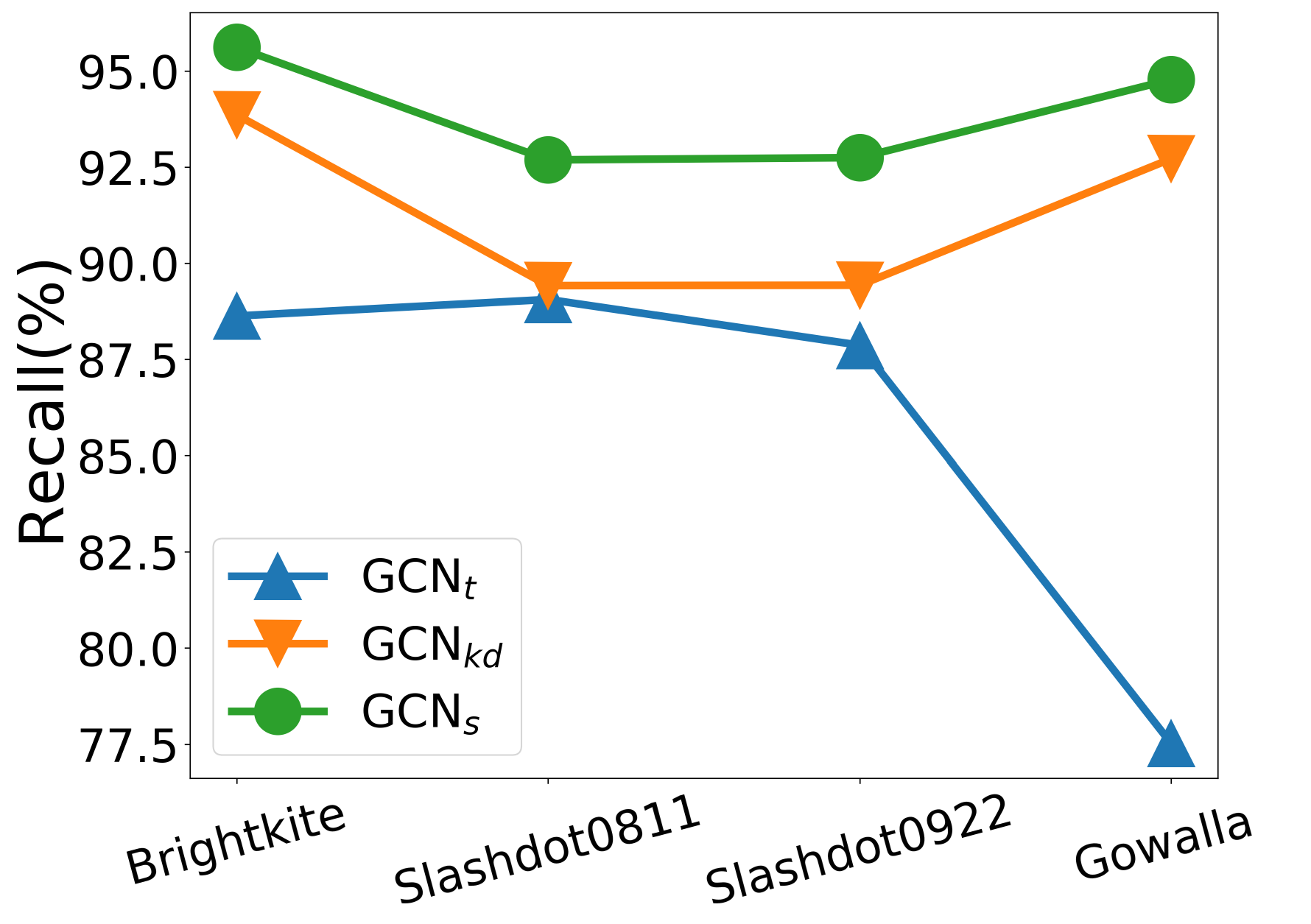}
		\caption{MIS}
		\label{subfig:recallmisreal2}
	\end{subfigure}
	\caption{Recall on synthetic and real-world datasets. GCN$_{s}$ always achieves higher recall than both GCN$_{t}$ and GCN$_{kd}$.}
	\label{fig:recall}
\end{figure}

Figure \ref{fig:recall} shows the recall of GCN models with different modules, where GCN$_{t}$ is the teacher GCN without KD and boosting, GCN$_{kd}$ is the student GCN with only KD, and GCN$_{s}$ is the student GCN with both KD and the problem-specific boosting module. GCN$_{kd}$ achieves higher recall than GCN$_{t}$ in some cases (Figure \ref{subfig:recallmvcreal2} and \ref{subfig:recallmisreal2}) while it achieves recall close to GCN$_{t}$ (Figure \ref{subfig:recallmvcreal1} and \ref{subfig:recallmisreal1}) and even lower than GCN$_{t}$ (Figure \ref{subfig:recallmvcsynthetic} and \ref{subfig:recallmissynthetic}). In contrast, GCN$_{s}$ obtains the highest recall in all the cases. Moreover, Table \ref{tab:mvcresultreallp} and \ref{tab:misresultreal} indicate that solutions generated by {\namemodel} (using GCN$_{s}$) achieve higher coverage on the MVC problem and have larger solution sizes on the MIS problem than those generated by {\namemodel$_{pt}$} (using GCN$_{t}$). Based on all the above, we draw a conclusion that GCN with problem-specific boosting has better performance because it pays more attention to the misclassified and problem-related nodes.

\section{Conclusion}

In this paper, we have proposed a novel framework called {\namemodel} to improve the efficiency of existing CO algorithms on large graphs by precisely reducing the search space. We have applied a GNN-based model to identify the promising nodes to prune the search space. Afterwards, we have applied the traditional CO algorithms on the reduced search space to obtain the final solution. Moreover, {\namemodel} adopts a KD framework with our designed problem-specific boosting module for further improvement. The efficiency and efficacy of our proposed {\namemodel} have been demonstrated on both synthetic and real-world datasets across two graph CO problems. Traditional CO algorithms with {\namemodel} are at least 2 times faster than their original versions.

\section{Acknowledgments}

We thank the anonymous reviewers for their valuable and constructive comments. This work was supported partially by the National Natural Science Foundation of China (grant \# 62176184), and the Fundamental
Research Funds for the Central Universities of China.

\bibliography{aaai24}

\clearpage

\appendix
\section{Supplementary Material}

\subsection{Traditional CO Algorithms}

In this section, we discuss three traditional CO algorithms used in our paper in detail, i.e., linear programming (LP), greedy algorithm (GD), and local search (LS). We also present their versions with our proposed {\namemodel}.

\subsubsection{Linear Programming.}

Since LP on the MVC problem has been introduced in the previous section (Our Proposed Method: {\namemodel}), we only give the formulations of the MIS problem in this section.

For a given graph $\mathcal{G}=(\mathcal{V},\mathcal{E})$, each node $v\in\mathcal{V}$ is denoted as a binary variable $x_{v}$, where $x_{v}=1$ means node $v$ is included in the final solution and $x_{v}=0$ otherwise. Then the LP formulation of the MIS problem is as follows:
\begin{align}
	\label{eqn:mislpobj} \max&\sum_{v\in\mathcal{V}}x_{v} \\
	\label{eqn:mislpconstraint1} &x_{v}\in\{0,1\} \\
	\label{eqn:mislpconstraint2} &x_{v}+x_{u}\le1,(v,u)\in\mathcal{E}
\end{align}
where Equation \ref{eqn:mislpobj} is the objective of MIS that maximizes the number of nodes in the solution; Equation \ref{eqn:mislpconstraint1} means the value of each node variable is 0 or 1; Equation \ref{eqn:mislpconstraint2} makes sure that at most one node is included in the solution for each edge in the graph.

Above is the general formulation of LP on the MIS problem, now we give the formulation of LP with {\namemodel} on the reduced search space (denoted by $\mathcal{V}_{g}$):
\begin{align}
	\label{eqn:ourmislpobj} \max&\sum_{v\in\mathcal{V}_{g}}x_{v} \\
	\label{eqn:ourmislpconstraint1} &x_{v}\in\{0,1\} \\
	\label{eqn:ourmislpconstraint2} &x_{v}+x_{u}\le1,(v,u)\in\mathcal{E},v\in\mathcal{V}_{g}~\mbox{or}~u\in\mathcal{V}_{g}
\end{align}
where Equation \ref{eqn:ourmislpobj} is the modified objective that maximizes the number of \textit{good nodes} in the solution; Equation \ref{eqn:ourmislpconstraint1} means the value of each node variable is 0 or 1; Equation~\ref{eqn:ourmislpconstraint2} makes sure that at most one \textit{good node} is included in the solution for each edge in the graph.

\subsubsection{Greedy Algorithm.}

The pseudocodes of GD on the MVC and MIS problems are shown in Algorithm \ref{alg:mvcgreedy} and \ref{alg:misgreedy} respectively, where $\mbox{deg}(\cdot)$ is used to calculate the degree of one node. Given a graph $\mathcal{G}=(\mathcal{V},\mathcal{E})$, we iteratively add the best node $v^{*}$ to the solution until the terminal condition is reached. 

\begin{algorithm}[!htbp]
	\small
	\caption{Greedy Algorithm for MVC}
	\label{alg:mvcgreedy}
	\textbf{Input}: Graph $\mathcal{G}=(\mathcal{V},\mathcal{E})$, degree function $\mbox{deg}(\cdot)$ \\
	\textbf{Output}: Solution set $\mathcal{S}$
	\begin{algorithmic}[1] 
		\State $\mathcal{S}\leftarrow\emptyset$
		\While{$\mathcal{E}\ne\emptyset$}
		\State $v^{*}\leftarrow\arg\max_{v\in\mathcal{V}}\{\mbox{deg}(v)\}$ \Comment{Select best node}
		\State $\mathcal{S}\leftarrow\mathcal{S}\cup\{v^{*}\}$
		\State $\mathcal{V}\leftarrow\mathcal{V}\verb|\|\{v^{*}\},\mathcal{E}\leftarrow\mathcal{E}\verb|\|\{(v^{*},u)|(v^{*},u)\in\mathcal{E}\}$
		\EndWhile
	\end{algorithmic}
\end{algorithm}

For the MVC problem (see Algorithm \ref{alg:mvcgreedy}), we pick the node with the maximum degree from the search space ($\mathcal{V}$ for the general GD and $\mathcal{V}_{g}$ for GD with {\namemodel}) in each step (line 3) and add it to the current solution $\mathcal{S}$ (line 4). Afterwards, we remove $v^{*}$ from the node set $\mathcal{V}$ and remove the edges incident to $v^{*}$ from the edge set $\mathcal{E}$ (line 5). Such process (lines 3 to 5) is repeated until all the edges are covered.

Similarly, for the MIS problem (see Algorithm \ref{alg:misgreedy}), we add the node with the minimum degree (selected from $\mathcal{V}$ for the general GD and $\mathcal{V}_{g}$ for GD with {\namemodel}) to the current solution $\mathcal{S}$ in each iteration and remove $v^{*}$ and its neighbors from the search space $\mathcal{V}$ (line 3 to 5) until $\mathcal{V}$ is empty.

\begin{algorithm}[!htbp]
	\small
	\caption{Greedy Algorithm for MIS}
	\label{alg:misgreedy}
	\textbf{Input}: Graph $\mathcal{G}=(\mathcal{V},\mathcal{E})$, degree function $\mbox{deg}(\cdot)$ \\
	\textbf{Output}: Solution set $\mathcal{S}$
	\begin{algorithmic}[1] 
		\State $\mathcal{S}\leftarrow\emptyset$
		\While{$\mathcal{V}\ne\emptyset$}
		\State $v^{*}\leftarrow\arg\min_{v\in\mathcal{V}}\{\mbox{deg}(v)\}$ \Comment{Select best node}
		\State $\mathcal{S}\leftarrow\mathcal{S}\cup\{v^{*}\}$
		\State $\mathcal{V}\leftarrow\mathcal{V} \verb|\| (\{v^{*}\}\cup\{v|(v^{*},v)\in\mathcal{E}\})$
		\EndWhile
	\end{algorithmic}
\end{algorithm}

As mentioned in the Experimental Evaluation section, GD itself faces one challenge that the efficiency of GD relies on the nature of CO problems thus the improvement in efficiency is limited. In specific, during the solution generation process of the MVC and MIS problems, we need to pick the node with the maximum or minimum degree in each iteration. Then the time complexity of GD is $\mathcal{O}(d|\mathcal{S}||\mathcal{V}|)$, where $|\mathcal{S}|$ is the solution size, $|\mathcal{V}|$ is the size of the search space (node numbers) and $d$ is the average degree. It will be reduced to $\mathcal{O}(d|\mathcal{S}||\mathcal{V}_{g}|)$ by {\namemodel}, where $|\mathcal{V}_{g}|$ is the size of the reduced search space (number of \textit{good nodes}).

\subsubsection{Local Search.}

\begin{table*}[!htbp]
	\small
	\centering
	\begin{tabular}{l|lllll}
		\toprule
		\textbf{Method} & \textbf{BA5K} & \textbf{BA10K} & \textbf{BA20K} & \textbf{BA50K} & \textbf{BA100K} \\
		\midrule
		LP & 2727 & 5417 & 10882 & 27460 & 55590 \\
		LP+\namemodel$_{pt}$ & 2702 (99.25) & 5394 (99.36) & 10734 (99.27) & 26931 (99.34) & 53739 (99.29) \\
		LP+\namemodel & 2799 (\textbf{99.90}) & 5534 (\textbf{99.86}) & 11085 (\textbf{99.86}) & 27726 (\textbf{99.88}) & 55378 (\textbf{99.86}) \\
		\midrule
		GD & 2823 & 5607 & 11227 & 28077 & 56154 \\
		GD+\namemodel$_{pt}$ & 2713 (99.25) & 5420 (99.36) & 10776 (99.27) & 27045 (99.34) & 53947 (99.29) \\
		GD+\namemodel & 2815 (\textbf{99.90}) & 5578 (\textbf{99.86}) & 11147 (\textbf{99.86}) & 27918 (\textbf{99.88}) & 55752 (\textbf{99.86}) \\
		\midrule
		LS & 2958 & 5917 & 11847 & 29552 & N/A \\
		LS+\namemodel$_{pt}$ & 2702 (99.25) & 5394 (99.36) & 10734 (99.27) & 26931 (99.34) & N/A \\
		LS+\namemodel & 2802 (\textbf{99.90}) & 5536 (\textbf{99.86}) & 11089 (\textbf{99.86}) & 27730 (\textbf{99.88}) & N/A \\
		\bottomrule
	\end{tabular}
	\caption{MVC results (solution size with coverage of all edges in the parenthesis) on synthetic datasets. LS is not test on BA100k (at least 2 days) thus the results are N/A. Baseline+{\namemodel} always outperforms Baseline+{\namemodel$_{pt}$} in the coverage and shows best performance on LS.}
	\label{tab:mvcresultsynthetic}
\end{table*}

LS has two major steps to obtain a solution: (1) initial solution generation, and (2) neighborhood solution exploration. In the first step, one can generate an initial solution randomly or greedily. In the second step, local changes (e.g., remove one node or replace one node with two nodes) are applied to improve the initial solution and gain the neighborhood solution until no improvement is achieved. 

Given a graph $\mathcal{G}=(\mathcal{V},\mathcal{E})$, both steps above are performed on the whole search space (i.e., $\mathcal{V}$) for the general LS. {\namemodel} performs them on the reduced search space (i.e., $\mathcal{V}_{g}$). For both the MVC and MIS problems, we adopt the \textit{first improvement} strategy, i.e., once we have obtained a better neighborhood solution, we update the current solution by this solution and step into the next local search process. More details are as follows:

Firstly, we discuss the MVC problem (pseudocode is in Algorithm \ref{alg:mvcls}). For the general LS, we initialize the solution $\mathcal{S}$ by randomly picking an uncovered edge and adding both of its connected nodes until all edges are covered (For LS with {\namemodel}, we initialize the solution $\mathcal{S}$ just using the good node set $\mathcal{V}_{g}$). Afterwards, for each node in the solution $\mathcal{S}$, if all its neighboring nodes are in $\mathcal{S}$, we remove this node from $\mathcal{S}$ and step into the next search iteration. Such process is repeated until no such node exits in $\mathcal{S}$.

\begin{algorithm}[!htbp]
	\small
	\caption{Local Search for MVC}
	\label{alg:mvcls}
	\textbf{Input}: Graph $\mathcal{G}=(\mathcal{V},\mathcal{E})$, improvement flag $I=$ True \\
	\textbf{Output}: Solution set $\mathcal{S}$
	\begin{algorithmic}[1] 
		\State Initialize $\mathcal{S}$
		\While{$I=$ True}
		\For{$v\in\mathcal{S}$}
		\State $I\leftarrow$ True
		\For{$u\in\{u|(v,u)\in\mathcal{E}\}$}
		\If{$u\notin\mathcal{S}$}
		\State $I\leftarrow$ False
		\State \textbf{break}
		\EndIf 
		\EndFor
		\If{$I=$ True} \Comment{First improvement}
		\State $\mathcal{S}\leftarrow\mathcal{S} \verb|\| \{v\}$
		\State \textbf{break}
		\EndIf
		\EndFor
		\EndWhile
	\end{algorithmic}
\end{algorithm}

Then it comes to the MIS problem (pseudocode is in Algorithm \ref{alg:misls}). For the general LS, we initialize solution $\mathcal{S}$ by randomly adding a node to $\mathcal{S}$ and removing this node and all its neighbors from the graph until all the nodes are removed from the graph. Similarly, for LS with {\namemodel}, we initialize solution $\mathcal{S}$ by randomly adding a node to $\mathcal{S}$ and removing this node and all its neighbors from the good node set $\mathcal{V}_{g}$. Afterwards, we use the LS method introduced in \cite{andrade2012fast}. The main idea is to iteratively replace one node $v\in\mathcal{S}$ with two nodes $i$ and $j$ until no such node can be found, where $i\notin\mathcal{S}$ and $j\notin\mathcal{S}$ are \textit{one-tight} (exactly one of its neighbors is in $\mathcal{S}$) neighbors of $v$ and they are not adjacent to each other, i.e., $v$ is the only neighbor of $i$ and $j$ in the solution $\mathcal{S}$.

\subsection{Experiment Details}

In this section, we demonstrate more implementation details and experiment results.

\subsubsection{Network Configuration.}

For both the MVC and the MIS problems, the teacher model is a 4-layer GCN and the dimensions of all the hidden layers are set to 128. For the student models, we use a 4-layer GCN for the MVC problem and a 3-layer GCN for the MIS problem. The dimensions of all the hidden layers of both student models are set to 32.

\subsubsection{Training and Testing Details.}

{\namemodel} is trained and tested on a server with a dual-core Intel(R) Xeon(R) Gold 6226R CPU @ 2.90GHz, 256 GB memory, NVIDIA RTX 3090 GPU, and Ubuntu 18.04.6 LTS operating system.

For both synthetic and real-world training graphs (BA1k and Cora respectively), we randomly select 50\% of the nodes for training and the remaining 50\% nodes are used for validating. We use the node degrees as the input features. We train a teacher model for 500 epochs with a learning rate of 0.001 and a dropout rate of 0.5. For a student model, we train it for 1000 epochs with a learning rate of 0.0001 and a dropout rate of 0.5. The value of temperature parameter $T$ is set to 1 and that of the $\lambda$ parameter is set to 0.8.

\begin{algorithm}[!htbp]
	\small
	\caption{Local Search for MIS}
	\label{alg:misls}
	\textbf{Input}: Graph $\mathcal{G}=(\mathcal{V},\mathcal{E})$, improvement flag $I=$ True, one-tight flag $T$ \\
	\textbf{Output}: Solution set $\mathcal{S}$
	\begin{algorithmic}[1] 
		\State Initialize $\mathcal{S}$
		\While{$I=$ True}
		\For{$v\in\mathcal{S}$}
		\State $I\leftarrow$ False
		\State $\mathcal{N}(v)\leftarrow\{u|(u,v)\in\mathcal{E},u\notin\mathcal{S}\}$
		\If{$|\mathcal{N}(v)|\ge2$}
		\State $\mathcal{P}\leftarrow\{(i,j)|i\in\mathcal{N}(v),j\in\mathcal{N}(v)\verb|\|\{i\}\}$
		\For{$(i,j)\in\mathcal{P}$} \Comment{One-tight judgement}
		\State $T\leftarrow$ True
		\For{$m\in\{m|(m,i)\in\mathcal{E}\}$}
		\If{$m\ne v$ \textbf{and} $m\in\mathcal{S}$}
		\State $T\leftarrow$ False
		\State \textbf{break}
		\EndIf
		\EndFor
		\For{$n\in\{n|(n,j)\in\mathcal{E}\}$}
		\If{$n\ne v$ \textbf{and} $n\in\mathcal{S}$}
		\State $T\leftarrow$ False
		\State \textbf{break}
		\EndIf
		\EndFor
		\If{$T=$ True \textbf{and} $(i,j)\notin\mathcal{E}$}
		\State $\mathcal{S}\leftarrow\mathcal{S}\cup\{i,j\}$
		\State $\mathcal{S}\leftarrow\mathcal{S}\verb|\|\{v\}$
		\State $I\leftarrow$ True
		\State \textbf{break}
		\EndIf
		\EndFor
		\EndIf
		\If{$I=$ True} \Comment{First improvement}
		\State \textbf{break}
		\EndIf
		\EndFor
		\EndWhile
	\end{algorithmic}
\end{algorithm}

\begin{table*}[!htbp]
	\small
	\centering
	\begin{tabular}{l|ccccc}
		\toprule
		\textbf{Method} & \textbf{BA5K} & \textbf{BA10K} & \textbf{BA20K} & \textbf{BA50K} & \textbf{BA100K} \\
		\midrule
		LP & 2266 & 4569 & 9142 & 22682 & 44322 \\
		LP+\namemodel$_{pt}$ & 2071 & 4156 & 8344 & 20879 & 41529 \\
		LP+\namemodel & \textbf{2140} & \textbf{4330} & \textbf{8657} & \textbf{21663} & \textbf{42186} \\
		\midrule
		GD & 2253 & 4549 & 9067 & 22674 & 45291 \\
		GD+\namemodel$_{pt}$ & 2071 & 4153 & 8343 & 20875 & 41527 \\
		GD+\namemodel & \textbf{2139} & \textbf{4326} & \textbf{8657} & \textbf{21654} & \textbf{43176} \\
		\midrule
		LS & 2090 & 4172 & 8381 & 20973 & 41983 \\
		LS+\namemodel$_{pt}$ & 2070 & 4150 & 8325 & 20832 & 41451 \\
		LS+\namemodel & \textbf{2136} & \textbf{4319} & \textbf{8625} & \textbf{21598} & \textbf{43038} \\
		\bottomrule
	\end{tabular}
	\caption{MIS results (solution size) on synthetic datasets. Baseline+{\namemodel} outperforms Baseline+{\namemodel$_{pt}$}. LS+{\namemodel}and even generates better solutions than the LS.}
	\label{tab:misresultsynthetic}
\end{table*}

\subsubsection{Evaluation Details.}

As discussed in the previous section (Experimental Evaluation), {\namemodel} can not cover all the edges for the MVC problem in some cases. One possible reason is that GCN is sensitive to the node features, i.e., it pays more attention to the nodes with high degrees and overlooks some nodes with low degrees. However, nodes with low degrees can also be necessary to construct the final solution for the MVC problem. To overcome the above issue, we can calculate a confidence score for each node as the input feature, which reflects how likely each node is to be included in the solution set. We leave it for the future work.

\subsubsection{Results on Synthetic Datasets.}

Table \ref{tab:mvcresultsynthetic} shows the MVC results (solution size with coverage of all edges in the parenthesis) on synthetic datasets. For LS, it costs 48 hours to generate a solution on BA50k, thus we do not test our method on BA100k and the results are marked as N/A. Table \ref{tab:misresultsynthetic} shows the MIS results (solution size) on synthetic datasets. Similar to the case in real-world datasets, our method generates high-quality solutions on both the MVC and MIS problems, i.e., our proposed {\namemodel} is also efficacious on synthetic datasets.


\end{document}